\documentclass[lettersize,journal]{IEEEtran}

\usepackage{amsmath,amsfonts,amssymb}
\usepackage{array}
\usepackage{booktabs}
\usepackage[caption=false,font=footnotesize,labelfont=sf,textfont=sf]{subfig}
\usepackage{textcomp}
\usepackage{stfloats}
\usepackage{url}
\usepackage{graphicx}
\usepackage{cite}
\usepackage{balance}
\usepackage{multirow}
\usepackage{capt-of}

\newcommand{\trfdf}{TR-FDF}
\newcommand{\xfixed}{X_0}

\begin{document}

\title{Task-Relevant Feature-Dynamics Fidelity Enables Zero-Shot Sim-to-Real Transfer for Robotic Ultrasound Scanning}



\author{%
Yizhao Qian,
Jiayuan Luo,
Wanyi Zhu,
Yameng Zhang,
Max Q.-H. Meng \IEEEmembership{Fellow, IEEE},
Yixuan Yuan$^{*}$ \IEEEmembership{Senior Member, IEEE},
and Li Liu$^{*}$ \IEEEmembership{Member, IEEE}
\thanks{This work was supported by the National Natural Science Foundation of China (Grant 62473192), the Guangdong Regional Cooperation Fund (Grants 2025A1515140001 and 2024A151540132), and the Hong Kong Research Grants Council General Research Fund (Grant 14220622). Corresponding authors: Li Liu (e-mail: liuli@gbu.edu.cn) and Yixuan Yuan (e-mail: yxyuan@ee.cuhk.edu.hk).}%
\thanks{Yizhao Qian and Yixuan Yuan is with the Department of Electronic Engineering, The Chinese University of Hong Kong, Hong Kong SAR, China.}%
\thanks{Jiayuan Luo, Wanyi Zhu, and Li Liu are with the School of Advanced Engineering, Great Bay University, and the Dongguan Great Bay Institute for Advanced Study, Dongguan, China.}%
\thanks{Yameng Zhang is with the Department of Mechanical Engineering, The University of Hong Kong, Hong Kong SAR, China.}%
\thanks{Max Q.-H. Meng is with the Shenzhen Key Laboratory of Robotics Perception and Intelligence and the Department of Electronic Engineering, Southern University of Science and Technology, Shenzhen, China. He is also Professor Emeritus in the Department of Electronic Engineering, The Chinese University of Hong Kong, Hong Kong SAR, China.}%
}

\markboth{Robotic Ultrasound Sim-to-Real Transfer}{Task-Relevant Feature-Dynamics Fidelity for Robotic Ultrasound}

\maketitle
\begin{abstract}
Robotic ultrasound policies operating directly on B-mode images require extensive interaction data, whereas real-robot data acquisition is costly and safety-constrained. Simulation offers a scalable alternative, but zero-shot transfer of raw-B-mode closed-loop policies depends not only on frame realism but also on whether simulated observations reproduce task-relevant changes induced by probe motion. A mismatch in this pose-observation relationship can invalidate the feedback behavior learned in simulation. We term the cross-domain consistency of such motion-induced feature changes task-relevant feature-dynamics fidelity (\trfdf) and study it as a property of the observation model.
A local contraction analysis shows that greater local sensitivity of \trfdf\ mismatch to probe motion reduces the local contraction margin, irrespective of the specific observation modality. This analysis motivates a \trfdf-oriented robotic ultrasound simulator. The simulator updates the probe pose according to policy actions and generates the corresponding structural masks. A structural intermediate domain shared by anatomical masks and ultrasound images constrains pose-driven feature evolution and reduces this sensitivity. Trajectory-level fixed noise suppresses pose-independent sampling jitter, while a low-step conditional flow model provides realistic B-mode observations in real time.
In phantom experiments on a robotic ultrasound platform, a policy trained exclusively in simulation succeeded in 390 of 400 zero-shot deployments across four target planes. The simulator maintained competitive single-frame image quality and generated observations at 67.1 Hz. Controlled interventions revealed a clear divergence between simulator-side and real-robot success as this sensitivity increased.
Ablation studies and comparisons with external baselines further showed that the proposed design better preserves feature dynamics, image quality, and rollout efficiency than alternative generation strategies. These results indicate that, in the local closed-loop setting studied here, \trfdf\ captures a transfer property not measured by single-frame realism.
\end{abstract}


\section{Introduction}
\IEEEPARstart{I}{n} recent years, robotic ultrasound systems have evolved from relying on manual operation and fixed scanning protocols toward machine-learning-driven platforms for perception and closed-loop control. Previous studies have achieved target-plane acquisition \cite{Hua21}, anatomical-structure tracking \cite{Jia23}, and scan navigation \cite{Su24,Jia25}. In these tasks, control policies can estimate the current probe state from continuous B-mode ultrasound images and output motion commands for the robotic arm \cite{Jia23,Jia25,Li21,Li21b}. However, learning observation-action relationships from high-dimensional images typically requires a large number of interaction episodes \cite{Has20,Bi22}. Reinforcement learning on patients or real robots is constrained by safety and practical feasibility, while expert demonstrations and high-quality annotations are difficult to collect at scale \cite{Has20,Li21,Li21b,Bi22,Hu24}. Therefore, training image-based control policies in simulation and subsequently deploying them on real robots provides an important means of reducing the need for real-world interaction and annotation.

This study considers zero-shot sim-to-real transfer in the strict sense: the policy is trained exclusively on ultrasound observations generated in simulation, and deployment on the real robot uses no real-robot interaction, online adaptation, or policy fine-tuning. Reliable real-robot closed-loop deployment under this setting remains unestablished. Accordingly, this paper asks: what properties must an ultrasound observation generator possess to enable a raw-B-mode policy trained in simulation to maintain reliable closed-loop behavior on a real robot?

In existing studies of autonomous robotic ultrasound servoing, dependence on real-domain information mainly takes two forms. One line of work introduces vessel segmentation, anatomical masks, or other structured intermediate representations at the perception front end, allowing the policy to learn control through relatively regular interfaces. These methods can achieve good autonomous scanning performance \cite{Bi22,Yan20,Vel22,Vel23}; however, their policy inputs are no longer raw B-mode images, and the perception process that maps raw images to structured interfaces also typically relies on real ultrasound data for training, calibration, or validation. Therefore, these methods fall outside the scope of this study. Another line of work retains raw B-mode images as the policy input and typically combines simulation frameworks with image-to-image generation to construct the observations required for policy training \cite{Hu24,Tom21,Vit25}. Because paired samples acquired from the same anatomy, at the same probe pose, and under the same imaging conditions are difficult to obtain for real and simulated ultrasound \cite{Tom21,Vit20,Zhu17}, these generators commonly use unpaired image translation, with single-frame appearance similarity, content preservation, or domain-distribution alignment as their primary optimization objectives \cite{Dom24,Fre25}. This approach can improve the realism of generated images and therefore comes closer to directly enabling raw-B-mode zero-shot transfer. However, visual similarity between single frames from the two domains does not guarantee that task-relevant features, such as vessels and anatomical boundaries, undergo consistent changes under the same probe motion \cite{Wan18,Par19}. Existing systems therefore often still require real-robot fine-tuning after simulation training \cite{Jam18}. This raises an unresolved question: beyond single-frame realism, what constraints must simulated ultrasound observations satisfy to support stable policy transfer to a real robot?

\begin{figure*}[!t]
\centering
\includegraphics[width=\textwidth]{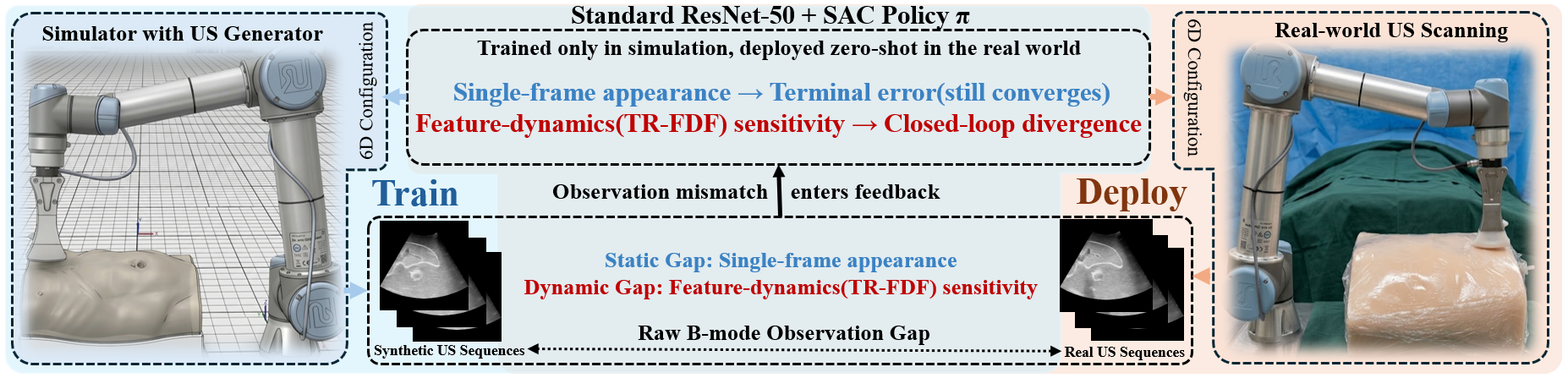}
\caption{Problem framework and the sim-to-real gap for raw B-mode observations. We identify TR-FDF sensitivity as a complementary metric beyond single-frame realism: the same probe motion should induce consistent task-relevant feature changes in simulation and on the real system. Motion-independent feature-transition mismatch and static residuals mainly enlarge the residual-error region around the target, whereas the local sensitivity of TR-FDF mismatch to probe motion reduces the effective contraction margin of the real closed loop.}
\label{fig:overview}
\end{figure*}

To answer this question, a simulator must reproduce not only realistic B-mode frames but also task-relevant feature changes induced by probe motion. Because each policy input follows a preceding probe motion \cite{Jia23,Li21}, discrepancies in the resulting changes between simulation and real ultrasound can lead to an incorrect observation-action relationship. Owing to stochastic ultrasound variations such as speckle, this requirement concerns statistical feature evolution rather than pixel-wise agreement. We refer to this observation-model property as task-relevant feature-dynamics fidelity (TR-FDF). Local closed-loop analysis (see Fig.~\ref{fig:overview}) shows that greater local sensitivity of TR-FDF mismatch to probe motion reduces the effective contraction margin, whereas static target bias and motion-independent perturbations mainly enlarge the ultimate error neighborhood. This distinction is independent of the observation modality. Thus, FID and simulator-side training success alone cannot guarantee stable zero-shot transfer \cite{Liu24m}.

Building on this analysis, we develop an ultrasound observation generator that jointly targets task-relevant feature evolution, single-frame realism, and rollout efficiency. Given a policy-generated probe pose, a structural renderer first produces multichannel anatomical masks. Stage I maps the masks and ultrasound images into a shared structural intermediate domain, using structural, cross-domain, and local-geometric constraints to preserve pose-driven anatomical changes while suppressing modality-specific factors. Stage II conditionally renders this representation into realistic ultrasound observations, with a frozen ultrasound encoder constraining their structural expression. Fixing the initial noise \(\xfixed\) along each trajectory suppresses pose-independent frame jitter, while Stage III uses flow reflow and few-step inference to enable real-time rollouts. Together, these components address observation-dynamics mismatch, single-frame realism, and rollout efficiency. To the best of our knowledge, this is the first demonstration of reliable zero-shot sim-to-real transfer for a policy trained exclusively on simulated raw B-mode observations under this strict setting. The contributions of this paper are as follows:

\begin{itemize}
    \item We propose task-relevant feature-dynamics fidelity (TR-FDF) to describe the consistency between simulated and real-world feature responses to the same probe motion. Local closed-loop analysis shows that the local sensitivity of TR-FDF mismatch to probe motion reduces the effective contraction margin, whereas static bias and motion-independent perturbations mainly enlarge the ultimate error neighborhood. This conclusion also applies to other observation modalities.
    \item We propose a robotic ultrasound simulator for raw-B-mode policy training. Its ultrasound observation generator combines a shared structural intermediate domain, trajectory-level fixed-$\xfixed$ conditional rendering, and few-step flow generation to jointly preserve pose-driven feature evolution, single-frame image realism, and rollout efficiency.
    \item On a real robotic ultrasound phantom platform, we verify that a raw-B-mode policy trained purely in simulation can be deployed directly on a real robot in a zero-shot manner. Controlled TR-FDF interventions, component ablations, and comparisons with external baselines further show that neither single-frame realism nor simulator-side policy success is sufficient to predict real closed-loop transfer. Under the setting studied in this paper, TR-FDF is an important property closely related to the transferability of the observation model.
\end{itemize}

\section{Related Work}
To delineate the research gap addressed in this paper, we first review existing characterizations of observation mismatch in closed-loop sim-to-real transfer and then discuss approaches for generating raw B-mode policy observations in robotic ultrasound.

\subsection{Observation Mismatch in Sim-to-Real Feedback Control}
In simulation-based closed-loop policy training, actions update the robot and environment states, from which the simulator generates the next observation. In conventional physics-based robotic simulators such as MuJoCo, explicit dynamics, scene geometry, and sensor models propagate joint states, velocities, and object poses into camera images, depth maps, and other observations \cite{Tod12,Mur21}. State or pose changes therefore have an explicit and traceable effect on observations, while motion-induced changes are constrained by dynamic and geometric relationships. This provides a structural anchor for state-observation evolution within the simulator.

For modalities whose observations cannot be directly rendered from system states and scene geometry, this motion-observation relationship is not guaranteed \cite{Zha22h,Aga20,Nar21,Ngo21}. In ultrasound, probe pose changes the spatial relationship between the probe and anatomy, while the resulting B-mode image also depends on acoustic propagation, scattering, attenuation, acoustic windows, probe and device responses, and post-processing \cite{Sha08,Sol25}. Thus, even at a fixed probe pose, inter-frame observations may contain speckle and artifact variations that are consistent with ultrasound imaging mechanisms but cannot be explained by pose changes alone \cite{Sha08,Sol25,Jud25}. Because paired data under identical anatomy, pose, and imaging conditions are difficult to obtain for simulated and real ultrasound \cite{Zhu17,Tom21,Vit20}, unpaired learned generators may learn a pose-observation relationship different from that of the real system or introduce appearance changes unrelated to probe motion \cite{Par19,Riv21}. This raises a central question: can such motion-induced observation mismatch affect closed-loop feedback and sim-to-real stability?

Existing theories of perception-based and partially observable control show that observation mismatch can affect closed-loop robustness and performance. Prior work on perception-based control has related state-dependent perceptual mismatch to closed-loop robustness and convergence \cite{Dea19}. Related work has analyzed performance loss through the closed-loop response \cite{Dea20b} and the effects of additive perceptual errors on state-estimation errors and tracking tubes \cite{Cho22b}. General contraction theory similarly shows that state-dependent disturbances may weaken effective contraction, whereas bounded additive disturbances mainly determine the ultimate error bound \cite{Dav21}. In partially observable stochastic control, changes in observation channels and measurement kernels propagate to filtering and control performance \cite{Yuk10,Dem25}; related studies consider state-, action-, and history-dependent observation perturbations \cite{Kra26}, as well as observation shifts in task-relevant representation spaces \cite{Mah23}. However, these studies and conventional simulator evaluations do not explicitly characterize the motion-conditioned task-feature transition mismatch that a sim-to-real observation generator must preserve to retain the feedback law learned in simulation. We therefore study task-feature transition distributions under identical state changes and define their cross-domain consistency as task-relevant feature-dynamics fidelity (TR-FDF). Extended-state stochastic Lyapunov analysis further shows that distributional mismatch increasing with motion changes the effective contraction factor of the real closed loop, whereas motion-independent jitter and static target bias mainly determine the ultimate error bound.

\subsection{Ultrasound Observation Generation for Robot Learning}
Existing ultrasound simulators fall into two categories. The first uses explicit acoustic or geometric forward models to generate B-mode images from anatomy, probe pose, and imaging parameters \cite{Jen04,Gar21,Sha08}. Based on scatterers, CT boundaries, or tissue acoustic properties, these models simulate impulse responses, reflection, attenuation, shadowing, and beam propagation \cite{Jen04,Sha08,Due25}, yielding an interpretable and structurally constrained pose-observation relationship \cite{Jen04,Gar21}. However, high-fidelity propagation is computationally expensive, whereas real-time approximations struggle to reproduce speckle, shadowing, tissue contrast, and device post-processing \cite{Wan20,Due25,Sol25}.

The second category uses learned simulators. Because sim-real B-mode pairs matched in pose, anatomy, and device conditions are difficult to obtain, these methods mainly rely on unpaired image translation \cite{Tom21,Vit20}. Adversarial, cycle-based, contrastive, or structure-preserving objectives improve brightness, texture, speckle, and device style, while content or segmentation constraints limit anatomical changes \cite{Tom21,Vit20,Son24}. However, unpaired mappings are underconstrained: realistic single frames need not preserve the correspondence between changes in input conditions and output observations \cite{Zhu17,Tom21,Son24}. Temporal discriminators, semantic consistency, and three-dimensional viewpoint constraints reduce flicker, semantic discontinuities, and long-term structural drift \cite{Par19,Riv21}, but visual coherence does not guarantee correct motion-conditioned responses. In particular, probe motion in real ultrasound may involve spatiotemporally structured speckle decorrelation \cite{Jud25}. Recent robotic ultrasound simulation work has likewise identified the lack of physics-based consistency between consecutive learned frames as a limitation \cite{Ao25}. Existing methods have not directly constrained or compared the conditional distributions of sim-real task-feature increments under the same probe motion. We argue that this motion-conditioned feature consistency, beyond static single-frame similarity, is critical for raw-B-mode zero-shot transfer. To the best of our knowledge, reliable zero-shot real-robot transfer of policies trained solely on simulated raw B-mode observations remains unreported across both simulator categories \cite{Ao25,Bi22}.

This gap motivates the TR-FDF formulation and closed-loop analysis presented in the next section.

\section{Problem Formulation and Analysis}
In this section, we address three questions: (1) which observation discrepancies affect sim-to-real transfer; (2) how these discrepancies affect local closed-loop error and stability margins; and (3) what properties a task-specific observation generator should satisfy.

\subsection{Task Setup and Notation}
\label{sec:formulation}
In closed-loop control based on B-mode images, a probe motion not only changes the current observation but also affects subsequent states through the next action. The causal relationship considered in this paper can be summarized as
\begin{equation}
\Delta p_{t-1}^e
\longrightarrow
\Delta y_{t-1}^e
\longrightarrow
y_t^e
\longrightarrow
u_t^e
\longrightarrow
s_{t+1}^e.
\end{equation}
Here, \(e\in\{\mathrm{sim},\mathrm{real}\}\) denotes the domain. The preceding probe motion \(\Delta p_{t-1}^e\) induces the task-relevant feature change \(\Delta y_{t-1}^e\), yielding the current feature \(y_t^e\). The policy then outputs \(u_t^e=\pi(y_t^e)\) and drives the system to the next state \(s_{t+1}^e\). Thus, the suitability of simulated observations for closed-loop transfer depends not only on single-frame realism, but also on whether the same probe motion produces consistent task-relevant feature changes in simulation and in the real system.

Let \(p_t^e\) denote the probe pose in domain \(e\), and let \(y_t^e=\Phi(x_t^e)\) denote the task-relevant feature extracted from the B-mode image \(x_t^e\). For a one-step probe motion and its corresponding feature change,
\begin{equation}
\Delta p_t^e=p_{t+1}^e-p_t^e,
\qquad
\Delta y_t^e=y_{t+1}^e-y_t^e.
\end{equation}
Because ultrasound imaging is stochastic, \(\Delta y_t^e\) is not deterministic even under the same \(\Delta p_t^e\). Therefore, under fixed local task conditions, we denote the conditional distribution of the task-feature increment by \(Q_e^{\Delta p}\):
\begin{equation}
\Delta y_t^e
\mid
\Delta p_t^e=\Delta p
\sim
Q_e^{\Delta p}.
\end{equation}

We refer to the consistency between \(Q_{\mathrm{sim}}^{\Delta p}\) and \(Q_{\mathrm{real}}^{\Delta p}\) as task-relevant feature-dynamics fidelity (TR-FDF), and define its mismatch as
\begin{equation}
D_{\mathrm{TR}}(\Delta p)
=
W_2\!\left(
Q_{\mathrm{sim}}^{\Delta p},
Q_{\mathrm{real}}^{\Delta p}
\right).
\end{equation}
A smaller \(D_{\mathrm{TR}}\) indicates higher TR-FDF. Unlike FID, which measures similarity between single-frame observation distributions, \(D_{\mathrm{TR}}\) measures whether the same probe motion produces consistent feature transitions in simulation and in the real system. The next subsection analyzes how this transition mismatch propagates into the error evolution of the real closed loop.

\subsection{How Simulator Bias Enters Transfer: A Local Perturbation Analysis}
\label{sec:tr_fdf_stability}

This subsection examines what happens when a policy that has converged near the target in the simulator closed loop is transferred to the real environment in the presence of TR-FDF mismatch. We distinguish the effects of TR-FDF mismatch from those of other residual factors on sim-to-real transfer.

Let \(s_t^e\) denote the system state in domain \(e\), and let \(V_t^e=V(s_t^e)\) denote a nonnegative task-error function, so that \(\sqrt{V_t^e}\) represents the probe's deviation from the target. The local comparison below evaluates the simulated and real transition kernels from the same current state, \(s_t^{\mathrm{sim}}=s_t^{\mathrm{real}}=s_t\), and thus \(V_t^{\mathrm{sim}}=V_t^{\mathrm{real}}=V_t\). A policy that converges in simulation typically satisfies
\begin{equation}
\left(
\mathbb E[V_{t+1}^{\mathrm{sim}}\mid s_t]
\right)^{1/2}
\le
\rho\sqrt{V_t},
\qquad 0<\rho<1.
\end{equation}
This condition indicates local contraction near the target in the stochastic Lyapunov sense. It allows different next-step errors from the same local state and even temporary error increases along individual trajectories, but requires the conditional root-mean-square error over possible outcomes to decrease overall. It is therefore more suitable than a deterministic Lyapunov condition for describing a stochastic ultrasound closed loop.

We assume that small probe motions induce only bounded sim-real dynamic discrepancies. Near the target, the TR-FDF mismatch satisfies
\begin{equation}
D_{\mathrm{TR}}(\Delta p_{t-1})
\le
d_0+L_{\mathrm{TR}}\|\Delta p_{t-1}\|,
\qquad
\|\Delta p_{t-1}\|\le K_p\sqrt{V_t}.
\end{equation}
Here, \(d_0\) bounds the mismatch that remains when the probe does not move, such as differences caused by speckle or generator noise. \(L_{\mathrm{TR}}\) describes how quickly the mismatch grows as the probe motion increases. The second bound states that the corrective motion becomes smaller as the probe approaches the target. These are local boundedness assumptions.

We make the following local regularity assumption: near the target, the policy action, robot dynamics, and Lyapunov function have local Lipschitz properties or finite local gains, and the resulting sim--real feature-transition mismatch can be propagated through the closed-loop chain to the policy action, the next state, and the task error. Under this assumption, there exist local constants $C,B\ge 0$ such that
\begin{equation}
\left\|
\sqrt{V_{t+1}^{\mathrm{real}}}
-
\sqrt{V_{t+1}^{\mathrm{sim}}}
\right\|_{2\mid s_t}
\le
C D_{\mathrm{TR}}(\Delta p_{t-1})+B,
\end{equation}
where
\begin{equation}
\|z\|_{2\mid s_t}
=
\left(\mathbb E[\|z\|^2\mid s_t]\right)^{1/2}.
\end{equation}
Here, \(C\) represents the closed-loop sensitivity to feature-transition mismatch, while \(B\) aggregates static feature bias, sim-real physical discrepancies, and other residual errors.

By the triangle inequality for conditional root-mean-square norms,
\begin{equation}
\left(
\mathbb E[ V_{t+1}^{\mathrm{real}}\mid s_t]
\right)^{1/2}
\le
\left(
\mathbb E[ V_{t+1}^{\mathrm{sim}}\mid s_t]
\right)^{1/2}
+C D_{\mathrm{TR}}(\Delta p_{t-1})+B.
\end{equation}
Substituting the simulator closed-loop contraction relation, the TR-FDF assumption, and the local motion bound gives
\begin{equation}
\begin{aligned}
\left(
\mathbb E[V_{t+1}^{\mathrm{real}}\mid s_t]
\right)^{1/2}
&\le
\rho\sqrt{V_t}
+C\bigl(d_0+L_{\mathrm{TR}}\|\Delta p_{t-1}\|\bigr)+B \\
&\le
\bigl(\rho+C K_pL_{\mathrm{TR}}\bigr)\sqrt{V_t}
+Cd_0+B.
\end{aligned}
\end{equation}
Define
\begin{equation}
\rho_{\mathrm{real}}=\rho+C K_pL_{\mathrm{TR}},
\qquad
\beta=Cd_0+B.
\end{equation}
The real closed loop then satisfies
\begin{equation}
\left(
\mathbb E[V_{t+1}^{\mathrm{real}}\mid s_t]
\right)^{1/2}
\le
\rho_{\mathrm{real}}\sqrt{V_t}+\beta.
\end{equation}

This inequality distinguishes the effects of three types of sim-real mismatch on real closed-loop transfer. When \(\rho_{\mathrm{real}}<1\), the real closed loop retains local contraction in the mean-square sense and contracts into a residual error neighborhood determined by \(\beta\). When \(\rho_{\mathrm{real}}\ge 1\), convergence of the simulator closed loop alone cannot guarantee convergence in the real environment. Here, \(L_{\mathrm{TR}}\) characterizes the local sensitivity of TR-FDF mismatch to probe-motion magnitude. Because it enters \(\rho_{\mathrm{real}}\), \textbf{increasing \(L_{\mathrm{TR}}\) directly consumes the contraction margin of the simulator closed loop.} The quantity \(d_0\) is the upper bound on feature-increment mismatch when the probe is stationary and mainly reflects feature jitter caused by differences in speckle or generation noise. The term \(B\) aggregates other residual effects not described by TR-FDF, including static task-feature bias introduced by the generator. FID can serve as an empirical indicator of related single-frame distribution differences, but it is not strictly equivalent to this bias. The terms \(d_0\) and \(B\) jointly enter the additive term \(\beta\), mainly producing residual error near the target rather than directly changing the contraction factor.

\begin{figure}[!t]
\centering
\includegraphics[width=\columnwidth]{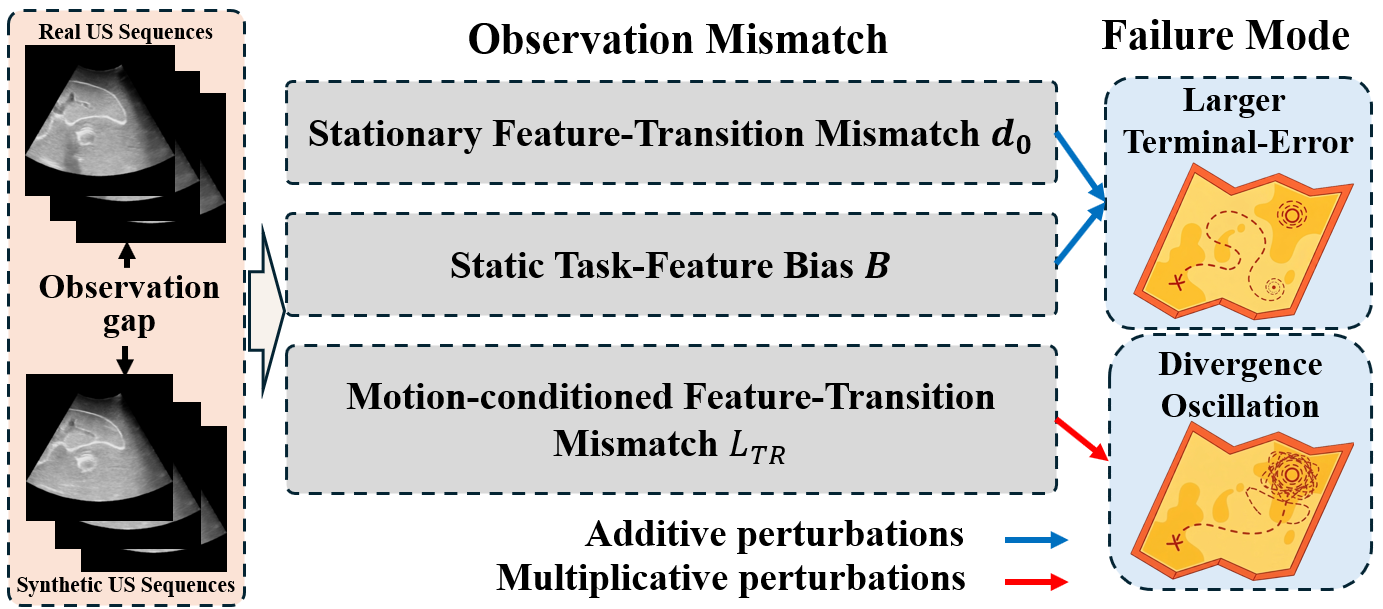}
\caption{Two effects of raw B-mode observation mismatch on closed-loop transfer. Feature-transition mismatch under a stationary probe, $d_0$, together with static task-feature bias and other residual terms, $B$, enters the additive term $\beta=Cd_0+B$ and mainly enlarges the residual-error region around the target. Under probe motion, the local TR-FDF sensitivity $L_{\mathrm{TR}}$ enters the effective contraction factor and reduces the contraction margin.}
\label{fig:observation_mismatch_effects}
\end{figure}

\subsection{Design Criteria for the Observation Generator}
\label{sec:criteria}
Section 3.2 shows that the observation generator can directly affect \(L_{\mathrm{TR}}\), \(d_0\), and the generator-induced static bias contained in \(B\), but cannot eliminate physical residuals such as robot dynamics. Accordingly, we summarize the requirements for the observation generator as three criteria.

\textbf{Criterion 1: TR-FDF.} \(L_{\mathrm{TR}}\) directly enters the contraction factor of the real closed loop. It characterizes the local sensitivity of sim-real feature-transition mismatch to probe-motion magnitude, that is, whether task features in simulation and in the real system exhibit consistent local responses to probe motion. In our analysis, TR-FDF denotes the cross-domain consistency of sim-real feature transitions, \(D_{\mathrm{TR}}(\Delta p)\) denotes the distributional mismatch under a given probe motion, and \(L_{\mathrm{TR}}\) denotes the local sensitivity of this mismatch to motion magnitude. Because \(L_{\mathrm{TR}}\) directly enters the contraction factor of the real closed loop, reducing it is the primary design objective of the observation generator. The generator must therefore preserve the correct response of task features to probe motion rather than merely produce visually continuous images, because smooth but incorrect feature changes can still destroy the contraction margin of the real closed loop.

\textbf{Criterion 2: Motion-Independent Observation Quality.} The additive term \(\beta=Cd_0+B\) describes residual effects that do not vanish with probe-motion magnitude. Here, \(d_0\) corresponds to feature-jitter mismatch when the probe is stationary and therefore requires generated observations to remain stable under identical imaging conditions. The generator-related component of \(B\) mainly corresponds to static task-feature bias and therefore requires the single-frame generation distribution to be close to that of real ultrasound. In the experiments, FID is used to evaluate related single-frame distribution differences, but it is not a direct measure of \(B\). Reducing these errors mainly shrinks the residual error near the target without directly changing the contraction factor.

\textbf{Criterion 3: Rollout Efficiency.} The observation generator must also satisfy the time budget of step-wise rollouts during policy training, which determines whether it can support large-scale closed-loop policy training.

In summary, the observation generator must jointly reduce motion-related TR-FDF mismatch \(L_{\mathrm{TR}}\), control the motion-independent additive observation error \(Cd_0+B\), and satisfy the required rollout inference speed. Section 4 addresses these criteria through shared structural representation, fixed-\(X_0\) trajectory rendering, conditional B-mode generation, and few-step flow inference.

\section{Methodology}
\subsection{Generator Architecture from Design Criteria}
\begin{figure*}[!t]
\centering
\includegraphics[width=\textwidth]{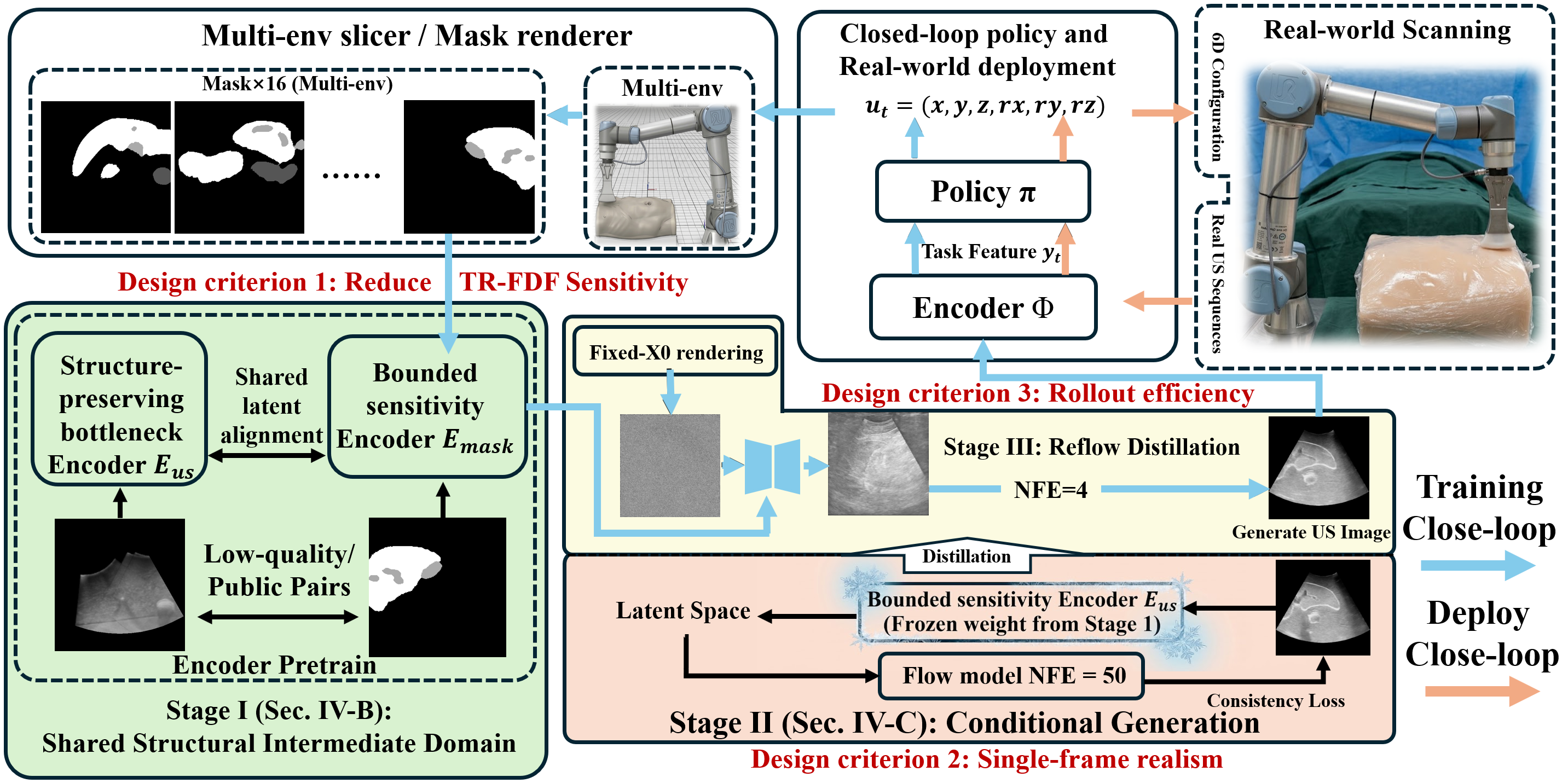}
\caption{Observation-generator architecture. Probe poses are converted into structural masks, encoded into a shared structural latent space in Stage I, rendered as B-mode observations by the CFM generator in Stage II, and accelerated by reflow distillation in Stage III. Fixed-$\xfixed$ trajectory rendering suppresses motion-independent sampling variation.}
\label{fig:system_overview}
\end{figure*}

Section~\ref{sec:criteria} summarizes three requirements for the observation generator: reducing $L_{\mathrm{TR}}$, the local sensitivity of TR-FDF mismatch to probe motion; controlling the motion-independent additive observation error $Cd_0+B$; and satisfying the rollout speed required for policy training.

To meet these requirements, we construct a three-stage B-mode observation generator and adopt fixed-$\xfixed$ rendering during trajectory generation. Given the probe pose $p_t$, the complete observation-generation and control pipeline is
\begin{equation}
p_t
\xrightarrow{R}
m_t
\xrightarrow{E_{\mathrm{mask}}}
z_t
\xrightarrow{\mathcal{G}(\cdot;\xfixed^*)}
\hat{x}_t
\xrightarrow{\Phi}
y_t
\xrightarrow{\pi}
u_t,
\end{equation}
where $R$ generates the structural mask $m_t$ from the probe pose. The Stage-I mask encoder maps $m_t$ to the shared structural latent $z_t$, and Stages II and III render this representation into the B-mode observation $\hat{x}_t$. The feature encoder $\Phi$ and policy $\pi$ then produce the control action $u_t$. The complete system is illustrated in Fig.~\ref{fig:system_overview} and follows three design principles.

\textbf{Design Principle 1: Reducing Motion-Related TR-FDF Mismatch.} The primary design objective is to reduce $L_{\mathrm{TR}}$, such that task-relevant features in simulation and in the real system exhibit similar local responses to probe motion. Stage I establishes the shared structural domain, and Stage II preserves the resulting structural changes in the generated B-mode images. Together, these stages reduce motion-related TR-FDF mismatch.

\textbf{Design Principle 2: Motion-Independent Observation Quality.} Fixed-$\xfixed$ rendering suppresses feature jitter caused by frame-wise random sampling and primarily reduces $d_0$. Stage II generates realistic B-mode appearance and primarily reduces the generator-induced component of the static observation bias contained in $B$. Together, these components control the additive observation error $Cd_0+B$.

\textbf{Design Principle 3: Rollout Efficiency.} Stage III uses reflow distillation to compress a high-step flow teacher into a low-step student. It brings the generation speed within the rollout time budget of policy training while preserving, to the extent possible, the structural responses and single-frame quality.

In summary, Stage I and structure-conditioned Stage II primarily reduce $L_{\mathrm{TR}}$. Fixed-$\xfixed$ rendering and conditional B-mode generation mainly control $d_0$ and the generator-related component of $B$, respectively, whereas Stage III satisfies the rollout-efficiency requirement. These modules do not map strictly one-to-one to the three criteria, but they address the principal error sources and engineering constraint identified in Section~\ref{sec:criteria}.

\subsection{Stage I: Shared Structural Domain}
\subsubsection{Design Objectives}
Stage I establishes the shared structural intermediate domain shown in Fig.~\ref{fig:system_overview}. It provides the B-mode renderer with a structural trajectory that varies continuously with probe pose while remaining invariant to stochastic ultrasound appearance. Because TR-FDF mismatch arises from discrepancies in pose-dependent task-feature changes, Stage I must preserve task-relevant, pose-driven anatomical dynamics in the latent passed to Stage II. We therefore learn a shared structural latent \(z_t\) that preserves structure, remains invariant to appearance, and exhibits bounded local responses.

Masks and ultrasound images represent the same anatomy but contain different information. Masks directly encode geometric boundaries and tissue labels, whereas ultrasound also contains modality-private appearance factors such as speckle, brightness, acoustic-window effects, and device post-processing. We express a real ultrasound observation as \(x_t=\Psi(m_t,a_t)\), where \(m_t\) denotes the pose-dependent structural condition and \(a_t\) denotes appearance factors that are independent of or weakly related to pose. Directly learning \(m_t\rightarrow\hat{x}_t\) can introduce random variations in \(a_t\) into inter-frame differences. Stage I instead learns a structural latent space shared by the mask and ultrasound domains, separating pose-driven structure from modality-private appearance.

\subsubsection{Network Architecture}
Specifically, Stage I comprises two encoders with identical architectures but independent parameters, \(E_{\mathrm{mask}}\) and \(E_{\mathrm{us}}\), together with a segmentation decoder \(D_{\mathrm{seg}}\). Given a rendered mask \(m_t\) and an ultrasound image \(x_t\), the two encoders produce
\begin{equation}
z_m=E_{\mathrm{mask}}(m_t),\qquad
z_u=E_{\mathrm{us}}(x_t).
\end{equation}
The encoders perform different functions but map their inputs to the same shared structural latent domain. \(E_{\mathrm{mask}}\) maps pose-conditioned masks to spatial latents that vary continuously with structure. Meanwhile, \(E_{\mathrm{us}}\) maps ultrasound images to the same structural domain while suppressing modality-private appearance factors such as speckle, brightness, and device style. The input resolution is \(224\times224\). After two stride-2 downsampling operations, both branches produce \(z\in\mathbb{R}^{8\times56\times56}\). Preserving a spatial resolution of \(56\times56\) allows the latent to retain local geometric layouts, including vessels, organ boundaries, and acoustic-window structures, rather than compressing them into a global semantic vector.

\subsubsection{Training Objectives and Loss Functions}
The Stage-I objectives are organized into two functional groups.

The first group preserves TR-FDF-related latent dynamics. \(\mathcal{L}_{\mathrm{dist}}\) matches the relative geometry of the mask and ultrasound latent spaces: sample pairs with larger pose-induced structural differences should remain farther apart, whereas pairs with smaller differences should remain closer. It therefore preserves the motion-dependent response magnitude associated with \(L_{\mathrm{TR}}\). \(\mathcal{L}_{\mathrm{smooth}}\) complements this batch-level geometric constraint by enforcing local continuity, such that masks with similar anatomy are mapped to nearby latent representations. Thus, \(\mathcal{L}_{\mathrm{dist}}\) preserves the relative scale of latent changes, whereas \(\mathcal{L}_{\mathrm{smooth}}\) prevents small structural changes from producing abrupt latent jumps. Finally, \(\mathcal{L}_{\mathrm{invar}}\) keeps \(E_{\mathrm{us}}\) stable under speckle, brightness, and gamma perturbations, suppressing pose-independent latent variation that primarily contributes to \(d_0\).

The second group prevents degenerate representations and establishes the shared structural space. \(\mathcal{L}_{\mathrm{seg}}\) requires both \(z_m\) and \(z_u\) to retain sufficient anatomical information to recover the corresponding segmentation labels, thereby directly preventing constant-latent collapse. \(\mathcal{L}_{\mathrm{align}}\) brings paired mask and ultrasound representations to the same structural location, while \(\mathcal{L}_{\mathrm{reg}}\) stabilizes the latent scale. Although alignment and regularization do not independently prevent collapse, together with segmentation supervision they produce a non-degenerate shared representation that preserves anatomical content across the two modalities. The overall training objective is
\begin{equation}
\begin{aligned}
\mathcal{L}_{\mathrm{Stage\,I}}
={}&
\underbrace{
\mathcal{L}_{\mathrm{dist}}
+\mathcal{L}_{\mathrm{smooth}}
+\mathcal{L}_{\mathrm{invar}}
}_{\text{TR-FDF dynamics}}
\\
&+
\underbrace{
\mathcal{L}_{\mathrm{seg}}
+\mathcal{L}_{\mathrm{align}}
+\mathcal{L}_{\mathrm{reg}}
}_{\text{non-collapse and shared structure}} .
\end{aligned}
\end{equation}

\subsubsection{Training Data}
Stage I is trained using a low-quality paired dataset and a small publicly available annotated dataset. The paired dataset is obtained by slicing 3D volumes reconstructed from ultrasound scans. It differs substantially from real ultrasound in sharpness, acoustic-window shape, and structural projection relationships, with an FID of 96.81. Stage I therefore uses these data to pretrain the shared structural domain rather than to learn the final realistic appearance. The small public annotated dataset supplements structural supervision and provides a basis for developing a more general Stage I in future work.

\subsubsection{Integration with Stage II}
After training, \(E_{\mathrm{mask}}\) converts pose-conditioned masks into structural latents for Stage II. \(E_{\mathrm{us}}\) remains frozen during Stage II and constrains generated images on the real-ultrasound side to remain within the same shared structural domain. This design constrains the realistic appearance rendering of Stage II through the structural representation learned in Stage I, rather than allowing it to learn an image-variation pathway unrelated to probe pose.

\subsection{Stage II/III: Structure-Conditioned B-mode Rendering and Efficient Distillation}
\subsubsection{Stage II Design}
Stage I outputs a structural latent \(z_t\) that varies with probe pose, whereas the downstream SAC policy requires raw B-mode images. Stage II therefore acts as a \textit{structure-conditioned appearance renderer}. It renders realistic ultrasound appearance over the bounded, non-collapsed structural trajectory provided by Stage I, primarily addressing the Motion-Independent Observation Quality criterion in Section~\ref{sec:criteria}. Stage II also has a limited but critical role in TR-FDF: its outputs must remain conditioned on \(z_t\), without amplifying, suppressing, or distorting structural changes between neighboring poses into unrelated image variations. Through structural conditioning and structural consistency, Stage II learns speckle, acoustic-window characteristics, gray-level distributions, and device style while transmitting changes in \(z_t\) to the image domain with bounded sensitivity, thereby preserving stable TR-FDF responses in the generated observations.

\subsubsection{Conditional B-mode Rendering}
We use conditional flow matching (CFM) as the Stage-II B-mode renderer. Given \(z_t\), the CFM model learns a conditional velocity field from initial noise \(x_0\) to a target B-mode image \(x_1\). Time embeddings are injected into every residual block, while \(z_t\) enters multiple resolution levels through spatially adaptive affine modulation:
\begin{equation}
    h \leftarrow h \cdot \bigl(1 + \gamma(z_t)\bigr) + \eta(z_t),
\end{equation}
where \(\gamma(z_t)\) and \(\eta(z_t)\) are generated from the Stage-I latent, keeping local texture, brightness, and boundary details conditioned on \(z_t\) across scales. Assuming that the velocity field is locally Lipschitz with respect to its conditioning input, the CFM flow map has a finite-gain response to \(z_t\). Stages I and II therefore jointly reduce \(L_{\mathrm{TR}}\).

\subsubsection{Real-Ultrasound Consistency Constraint}
Stage II is trained using unpaired real ultrasound images. For each randomly sampled real image \(x\), the frozen ultrasound encoder extracts its structural latent \(z_u=E_{\mathrm{us}}(x)\). The CFM model then reconstructs an image \(\hat{x}\) conditioned on \(z_u\), and a reconstruction-consistency loss minimizes the difference between \(\hat{x}\) and the source image \(x\). In this way, each real image provides its own condition--target pair, enabling the CFM model to learn real-ultrasound appearance without requiring paired simulated masks and real ultrasound images.

\subsubsection{Stage III: Reflow Distillation for Rollout Efficiency}
The high-NFE Stage-II teacher provides better image quality, but its inference speed is insufficient for large-scale closed-loop policy rollouts. Stage III compresses it into a low-step student through reflow distillation. Given teacher-generated pairs \((x_0,x_1)\), where \(x_1\) is the high-NFE terminal output, the student learns approximately straight paths:
\begin{equation}
\mathcal{L}_{\mathrm{reflow}}
=
\left\lVert
v(x_\tau, \tau, z_t) - (x_1-x_0)
\right\rVert_2^2 .
\end{equation}
The distilled student generates B-mode observations at NFE\(=4\), satisfying the rollout-efficiency requirement. Stage III targets inference speed rather than further improvement of TR-FDF and should preserve the teacher's appearance statistics and structure-conditioned responses as much as possible.

\subsection{Fixed-\(\xfixed\) Trajectory Rendering: Reducing Generator-Side Pose-Independent Observation Jitter}
\label{sec:fixed_x0}
Frame-wise independent sampling during standard flow-matching inference introduces pose-independent inter-frame jitter. Let \(\xi_t\sim\mathcal{N}(0,I)\) denote the independently sampled initial noise for frame \(t\), and let \(F_N(z_t,\xi_t)\) denote the terminal output of the \(N\)-step Euler flow map. Then
\begin{equation}
\begin{split}
    \hat{x}_{t+1} - \hat{x}_t
    ={}&
    \underbrace{
    F_N(z_{t+1}, \xi_{t+1})
    -
    F_N(z_t, \xi_{t+1})
    }_{\substack{\text{Structure-driven} \\ \text{term } (\Delta z_t)}}
    \\
    &+
    \underbrace{
    F_N(z_t, \xi_{t+1})
    -
    F_N(z_t, \xi_t)
    }_{\substack{\text{Generator-side} \\ \text{resampling jitter}}} .
\end{split}
\end{equation}
The first term is driven by \(\Delta z_t\) and represents the observation dynamics that the policy should learn during simulated rollouts. The second arises from frame-wise resampling and is independent of probe pose. After feature extraction, it can contribute to the generator-side component of \(d_0\) and does not vanish as \(\lVert\Delta p\rVert\rightarrow0\). We therefore sample \(\xi^\star\sim\mathcal{N}(0,I)\) once at the beginning of each trajectory and reuse it throughout:
\begin{equation}
    \hat{x}_{t+1} - \hat{x}_t
    =
    F_N(z_{t+1}, \xi^\star) - F_N(z_t, \xi^\star) .
\end{equation}
This operation changes neither the network weights nor the marginal single-frame noise distribution; it changes only the within-trajectory sampling correlation, so inter-frame changes during training rollouts arise primarily from probe motion and the Stage-I structural latent. Fixed-\(\xfixed\) rendering therefore targets the additive-error route identified in Section~III. By removing generator-side resampling jitter, it reduces the generator-induced contribution to \(d_0\), and hence decreases \(\beta=Cd_0+B\), without directly changing the contraction factor \(\rho_{\mathrm{real}}\). According to the bound in Section~III, its primary effect is therefore to reduce the residual error neighborhood near the target. The component-level ablation in Section~\ref{sec:component_ablation} further shows that random frame-level \(\xfixed\) leaves single-frame realism and simulator learnability nearly unchanged but increases the operational TR-FDF sensitivity proxy and degrades real-robot deployment. Because this proxy can respond to stochasticity under finite motion bins and samples, this increase should not be interpreted as an isolated increase in theoretical \(L_{\mathrm{TR}}\).

\section{Experiments}
\label{sec:experiments}
This section evaluates whether TR-FDF is a measurable observation-model property relevant to zero-shot transfer in raw B-mode robotic ultrasound. We first describe the experimental platform, evaluation protocol, and TR-FDF sensitivity proxy in Sections~V-A and~V-B. We then address five questions:

\begin{enumerate}
    \item Can the proposed simulator support multi-plane 6DoF zero-shot
    transfer on a real robot (Section~V-C)?

    \item Does higher TR-FDF sensitivity correspond to lower real-robot
    transfer success (Section~V-D)?

    \item Which components support TR-FDF, single-frame realism, and
    rollout efficiency (Section~V-E)?

    \item How does the proposed simulator compare with external
    observation-generation baselines (Section~V-F)?

    \item Does the trained policy remain robust to contact-force and
    target-motion disturbances (Section~V-G)?
\end{enumerate}

\subsection{Experimental Platform, Policy, and Evaluation Protocol}
\begin{figure}[!t]
\centering
\includegraphics[width=\columnwidth]{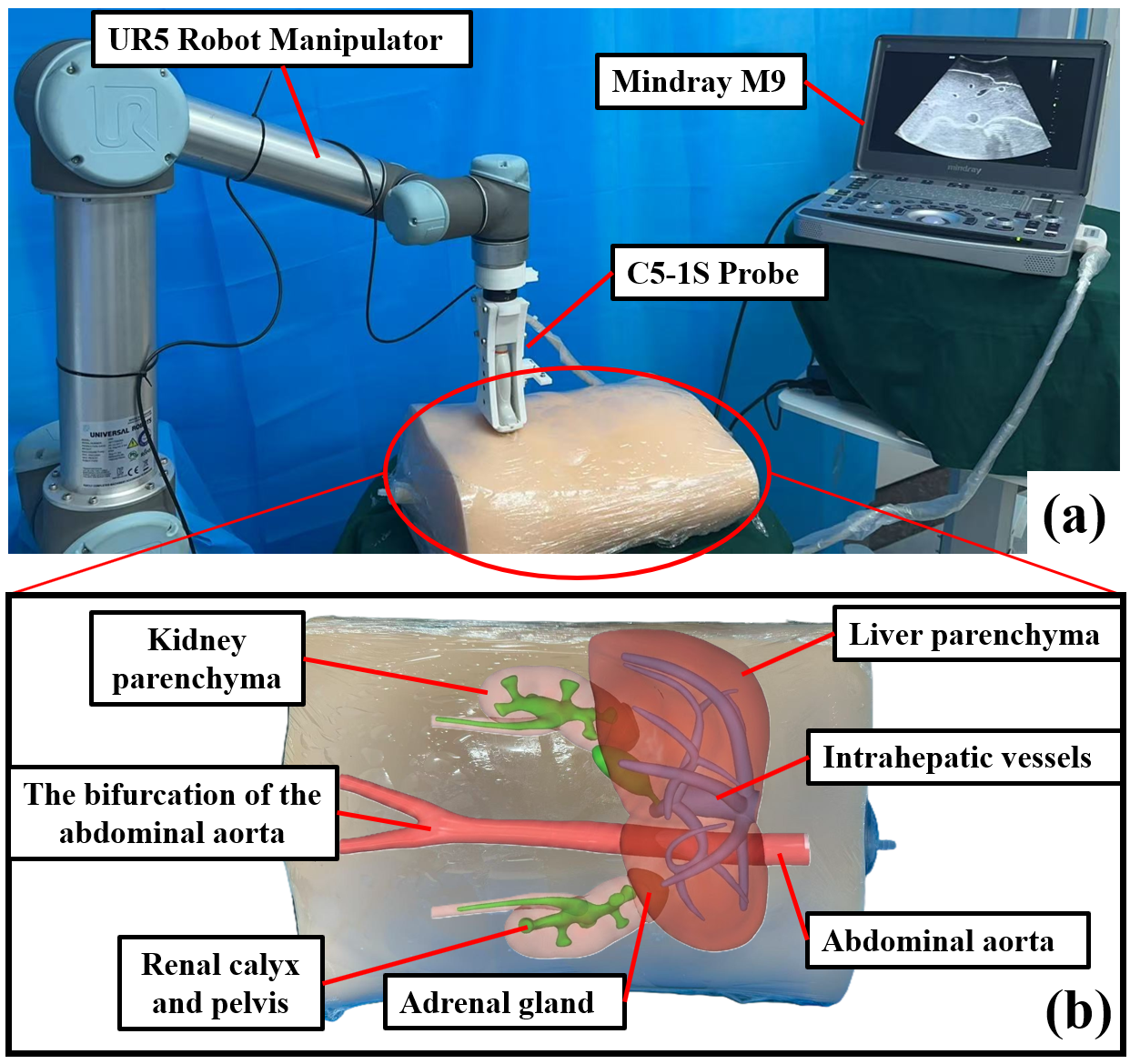}
\caption{Experimental platform and abdominal phantom. (a) Real-robot autonomous ultrasound platform, including the UR5 manipulator, Mindray M9 ultrasound scanner, C5-1S convex probe, force sensor, and custom abdominal phantom. (b) Internal layout of the custom abdominal phantom.}
\label{fig:platform}
\end{figure}

All real-robot experiments used a Universal Robots UR5 manipulator, a Mindray M9 ultrasound scanner, and a C5-1s convex probe operating at 1--6~MHz. The probe was mounted on the robot end effector. Real-time B-mode images were used as policy input, while robot pose and force-sensor readings provided the closed-loop control state. The main experiments were conducted on a custom abdominal phantom containing the liver, hepatic veins and arteries, kidneys, renal calyces, renal pelvis, and abdominal aorta (Fig.~\ref{fig:platform}). This phantom provided multiple anatomical structures, multiple target planes, and complex acoustic appearances, so the evaluation tested not only servo convergence to one view but also stable deployment across different anatomical conditions.

Generator training, reinforcement-learning training, and real deployment were performed on the same workstation with an Intel Core i9-14900K CPU and an NVIDIA RTX 5090 GPU. The training and inference system was implemented under Ubuntu 24.04 and ROS2. ROS2 handled robot communication, ultrasound image acquisition, force-sensor reading, policy inference, and low-level command publishing. In this paper, the pure-sim setting means that policy training used only simulated observations and simulated interactions: no real-robot images or real-robot interaction data entered policy training, and deployment did not use online fine-tuning, reward adaptation, or parameter updates.

The policy used a ResNet-50 visual front end, an MLP control head, and Soft Actor-Critic (SAC). This deliberately conventional and relatively simple architecture has limited generalization capacity compared with stronger task-specific visual or control models. We used it to test whether the proposed simulator can still support convergence and zero-shot transfer on the challenging high-dimensional servoing task, rather than relying on a highly specialized policy architecture. Unless otherwise stated, the task was 6DoF probe servoing: the image policy \(\pi\) controlled 5DoF motion, comprising two in-plane translations and three rotations, while a low-level force controller regulated the remaining axial contact DoF. This is a standard decomposition in robotic ultrasound: force control maintains stable and safe probe--surface contact, while the learned policy performs image-guided pose adjustment.

Each real-robot target or experimental condition used 100 independent episodes by default. An episode was counted as successful when the final position error was below \(3\,\mathrm{mm}\) and the final rotation error was below \(5^\circ\). This sample size gave several-percentage-point statistical resolution for binary success while keeping real robotic ultrasound platforms testing feasible. For a high target success rate \(p_0=0.97\), a 98\% confidence level, and an error radius \(\epsilon=0.04\), the normal approximation for Bernoulli trials gives
\begin{equation}
    n \ge \frac{z_{0.99}^2 p_0(1-p_0)}{\epsilon^2}
\end{equation}
where \(z_{0.99}=2.326\), giving \(n\approx99\). We therefore used 100 real-robot episodes as the default testing scale for each target or condition. In Section~V-C, four target planes were each tested for 100 episodes, yielding 400 episodes in total; this design supports per-plane diagnosis and gives a tighter overall success-rate estimate.

\subsection{Operational Proxy for TR-FDF Local Sensitivity}
\label{sec:metric}

Section~III defines $D_{\mathrm{TR}}$ as the distributional mismatch between simulated and real feature transitions under a given probe motion, and $L_{\mathrm{TR}}$ as the maximum local sensitivity of this mismatch to motion magnitude. Accordingly, the experimentally reported TR-FDF sensitivity serves as an empirical proxy for $L_{\mathrm{TR}}$. As shown in Section~III, $L_{\mathrm{TR}}$ enters the contraction factor of the real closed loop.

To extract task-relevant features, we use two frozen ResNet-50 encoders, denoted by $\Phi_{\mathrm{pose}}$ and $\Phi_{\mathrm{seg}}$, both trained exclusively on real ultrasound images. For simulated and real trajectories, we randomly sample pairs of frames and compute their feature increments. These increments are assigned to motion bins according to the probe-motion magnitude between the two frames. Within each bin, we estimate the simulated and real feature-transition distributions and compute the distributional distance between them.

We then treat the distributional distance in each motion bin as a function of motion magnitude. We fit this relationship and use finite differences to estimate its local slopes. The maximum estimate is reported as the $L_{\mathrm{TR}}$ proxy, which we refer to as TR-FDF sensitivity. The stationary or minimum-motion bin is used separately to estimate $d_0$.

According to the analysis in Section~III, a lower TR-FDF sensitivity indicates that sim-to-real feature-transition mismatch increases more slowly with probe motion and therefore consumes less of the real closed-loop contraction margin. This value is an empirical estimate of the theoretical $L_{\mathrm{TR}}$ under finite motion bins and finite samples, rather than the theoretical upper bound itself. FID is reported separately for single-frame observation quality and is not combined with the $L_{\mathrm{TR}}$ proxy.

\subsection{Zero-Shot Transfer in Multi-Plane 6DoF Servoing}
This study provided the paper's main real-robot validation. It tested whether a simulator built around raw B-mode observation generation could support multi-plane zero-shot transfer of a robotic ultrasound servoing policy on the real platform. Each target plane was tested with 100 independent real-robot trials, for 400 episodes in total. During deployment, the policy received the real B-mode image stream and output closed-loop probe motion commands; no real-robot interaction training or policy fine-tuning was used at deployment.

Each trial started from a randomized initial pose near the target plane. The initial position error ranged from a few millimeters to more than \(100\,\mathrm{mm}\), and the initial rotation error covered task-relevant perturbations around all three rotation axes. A trial was counted as successful when the robot end effector remained within \(3\,\mathrm{mm}\) translation error and \(5^\circ\) rotation error of the target. Failure was triggered by exceeding the scanning range, exceeding a \(20^\circ\) rotation limit, or exceeding the 60 s time limit. In Table~\ref{tab:expA_main}, out-of-bounds and timeout failure types correspond to pose-limit and time-limit termination, respectively.

The proposed ultrasound simulator with four-step inference achieved a TR-FDF sensitivity proxy of 31.61, an FID of 29.66, and an inference speed of 67.1 Hz, corresponding to \(14.9\,\mathrm{ms/frame}\). The policy-training observations were grayscale B-mode images with a resolution of \(224\times224\). The four target planes were selected at random and were located approximately over the hepatic vein, right kidney, abdominal aorta, and liver-tip regions.

\begin{figure*}[!t]
\centering
\includegraphics[width=\textwidth]{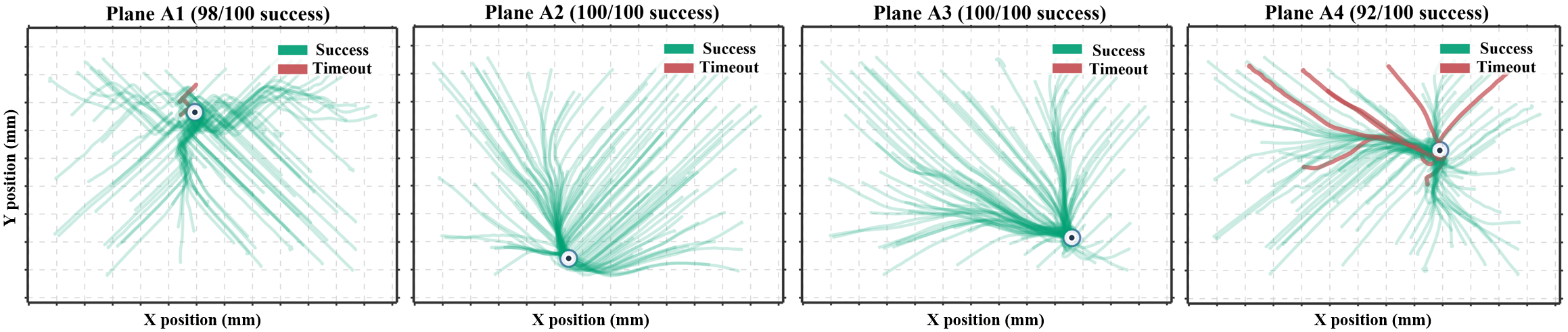}
\caption{Real-robot zero-shot convergence trajectories for four target planes. Each panel shows top-down XY trajectories on A1--A4, with 100 episodes per plane and 400 deployments in total. Green denotes successful episodes, and red denotes timeout failures; the center marker indicates the target position, and the surrounding region indicates the $3\,\mathrm{mm}$ translational success threshold.}
\label{fig:expA-traj}
\end{figure*}

\begin{figure}[!t]
\centering
\includegraphics[width=\columnwidth]{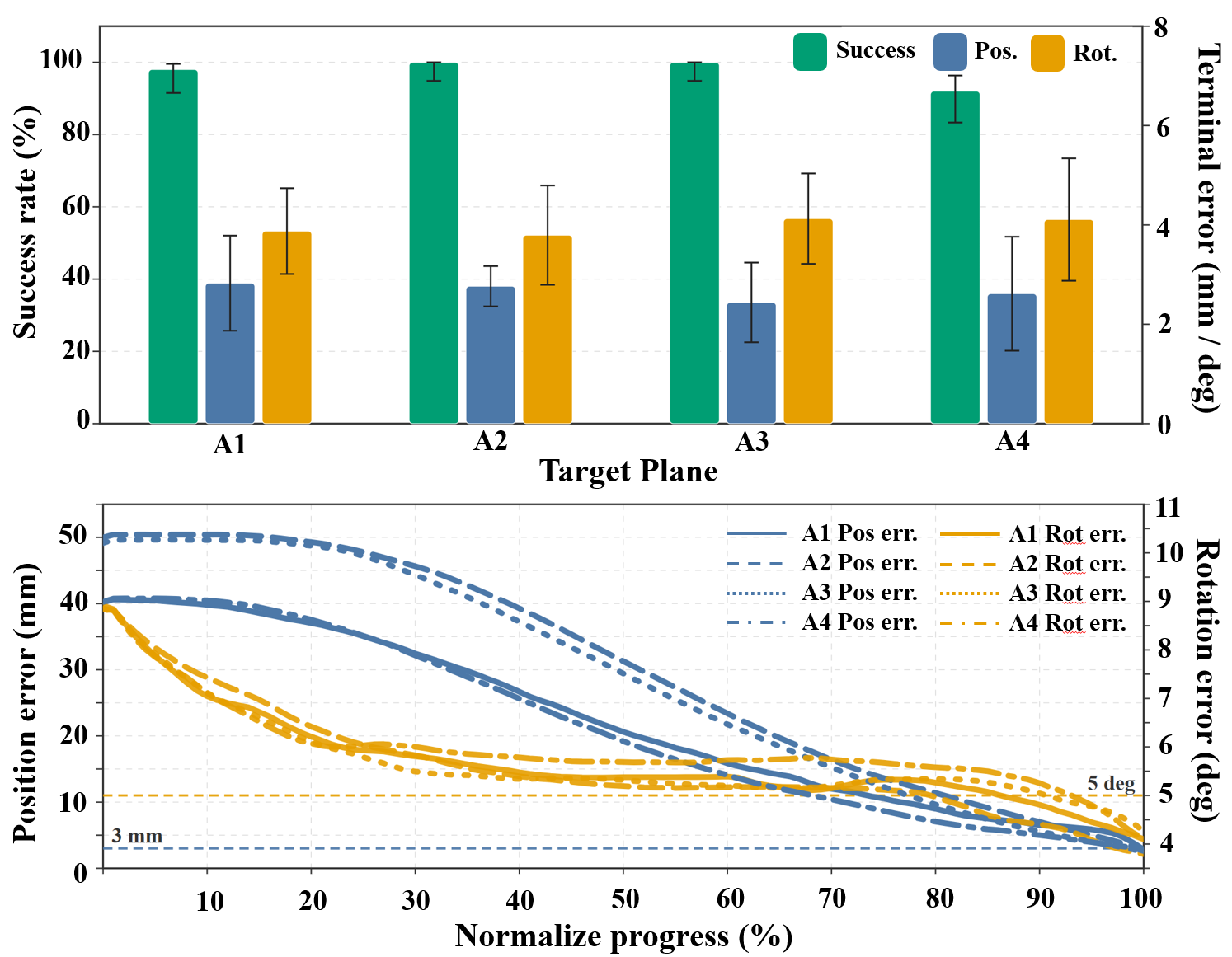}
\caption{Zero-shot deployment performance. (a) Success rate, terminal position error, and terminal rotation error for the four target planes, reported with means and standard deviations. (b) Changes in mean position and rotation errors.}
\label{fig:expA-summary}
\end{figure}

Table~\ref{tab:expA_main} summarizes the real-robot results across the four target planes. Overall, the policy succeeded in 390/400 real-robot episodes, giving an overall success rate of 97.5\%, and the mean episode duration for every plane was below 30~s. All failures were timeouts, and no out-of-bounds failure occurred. Fig.~\ref{fig:expA-traj} shows the top-down XY trajectories from all 400 real-robot deployments, and Fig.~\ref{fig:expA-summary} summarizes success rate, terminal accuracy, and step-wise convergence.

These results show that the proposed simulator can generate raw B-mode observations that support zero-shot deployment of a multi-plane 6DoF autonomous ultrasound servoing policy with a high real-world success rate. A2 and A3 reached 100/100 success, A1 reached 98/100, and A4 reached 92/100. The lower success rate on A4 may be related to weaker structural cues near the liver tip, which could limit visual feedback along some servoing directions. The results suggest potential for extending the framework to additional target planes and anatomical structures. The trajectory projections in Fig.~\ref{fig:expA-traj} and the convergence curves in Fig.~\ref{fig:expA-summary} are consistent with stable long-horizon observation--action coupling; the overall success rate was accompanied by a 98\% Wilson confidence.

\begin{table*}[!t]
\caption{Real-robot zero-shot deployment results for four target planes and overall results.}
\label{tab:expA_main}
\centering
\resizebox{\textwidth}{!}{%
\begin{tabular}{lcccccccccc}
\toprule
Plane & Trials & FID & TR-FDF sensitivity  & Speed & Success & 98\% CI & Pos. err. (mm) & Rot. err. (deg) & Duration (s) & Failure type \\
\midrule
A1 & \multirow{4}{*}{100} & \multirow{4}{*}{29.66} & \multirow{4}{*}{31.61} & \multirow{4}{*}{67.1 Hz} & 98/100 & 91.5--99.6 & \(2.83\pm0.95\) & \(3.88\pm0.86\) & \(28.52\pm11.14\) & 2 timeouts \\
A2 &  &  &  &  & 100/100 & 94.9--100.0 & \(2.77\pm0.40\) & \(3.79\pm0.99\) & \(23.79\pm8.38\) & No Failure \\
A3 &  &  &  &  & 100/100 & 94.9--100.0 & \(2.44\pm0.80\) & \(4.13\pm0.91\) & \(25.30\pm7.90\) & No Failure \\
A4 &  &  &  &  & 92/100 & 83.3--96.4 & \(2.61\pm1.14\) & \(4.11\pm1.23\) & \(28.62\pm11.48\) & 8 timeouts \\
Overall & 400 & 29.66 & 31.61 & 67.1 Hz & 390/400 & 95.0--98.8 & \(2.66\pm0.88\) & \(3.98\pm1.02\) & \(26.56\pm10.07\) & 10 timeouts \\
\bottomrule
\end{tabular}}
\end{table*}

\subsection{Controlled Interventions on TR-FDF Sensitivity}
This section examines whether real closed-loop convergence degraded as TR-FDF sensitivity increased, as predicted by the analysis in Section~III. The evaluation used two complementary interventions to test this prediction. The real-image lookup intervention in Section~V-D.1 perturbed a deterministic lookup simulator constructed from real B-mode images, providing controlled evidence with limited confounding. The learned-simulator structural intervention in Section~V-D.2 applied structural perturbations to the learned simulator and examined the same failure mode in the complete 6DoF pipeline. In both interventions, real-world success continuously decreased as TR-FDF sensitivity increased and remained lower than simulator success.

\subsubsection{Real-Image Lookup Intervention}
The real-image lookup intervention primarily perturbed the pose-to-observation relationship while controlling other factors as much as possible, including appearance realism and temporal noise flicker. This design directly established the relationship among TR-FDF sensitivity, simulator success, and real-world success.

To minimize interference from other factors, we did not directly perturb the learned generator. Generator-side interventions could simultaneously alter single-frame appearance, rendering stochasticity, and structural bias. These coupled changes would make transfer failures difficult to attribute specifically to the pose-to-observation relationship. We therefore constructed a real-image lookup simulator. It collected real ultrasound images within a plane and organized them into a lookup table, where each position stored an ultrasound image acquired in the real world. During policy training, the simulator determined the next position from the current robot state and action, then returned the corresponding ultrasound image as the policy input. We perturbed the queried positions to simulate changes in TR-FDF sensitivity and investigated how these perturbations affected the success rates.

At pose \(p\), an unperturbed query returned the corresponding real image \(y_{\mathrm{real}}(p)\). A perturbed query returned the real image associated with a displaced lookup position:
\begin{equation}
    y_d(p)=y_{\mathrm{real}}(p+\delta_d(p)),
\end{equation}
where \(d\) denotes the prescribed mean of the positional perturbation sampler. For each perturbation condition, the displacement field was sampled only once and then held spatially fixed. Consequently, the same pose always returned the same frame, thereby isolating the influence of \(d_0\) described in Section~III. In addition, the target lookup position was left unperturbed, satisfying \(\delta_d(p^*)=0\). The intervention therefore altered only the pose-to-observation relationship away from the target, without introducing frame-wise rendering noise or directly changing the target observation. All conditions shared the same real-image base, robot and force-control configuration, initialization distribution, and success criterion.

Table~\ref{tab:expB1_lookup} presents the results of this intervention experiment. Across all perturbation levels, FID remained within 0.628--1.100, indicating that image quality was unaffected. Meanwhile, TR-FDF sensitivity increased from 10.99 to 54.82, while real-world success decreased from 100\% to nearly 0\%. These results show a negative relationship between TR-FDF sensitivity and real-world success under comparable image quality.

Notably, simulator and real-world success became severely separated at intermediate perturbation magnitudes. At \(d=5.0\,\mathrm{mm}\), corresponding to a TR-FDF sensitivity of 18.09, simulator success remained 100/100, whereas real-world success decreased to 69/100. At \(d=10.0\,\mathrm{mm}\) and \(12.5\,\mathrm{mm}\), simulator success was 78/100 and 77/100, respectively, whereas real-world success was only 11/100 and 3/100. This finding explains why some policies achieve high success rates in simulation but remain difficult to transfer to real robots: this discrepancy is precisely the effect introduced by TR-FDF sensitivity.

\begin{figure}[!t]
\centering
\includegraphics[width=\columnwidth]{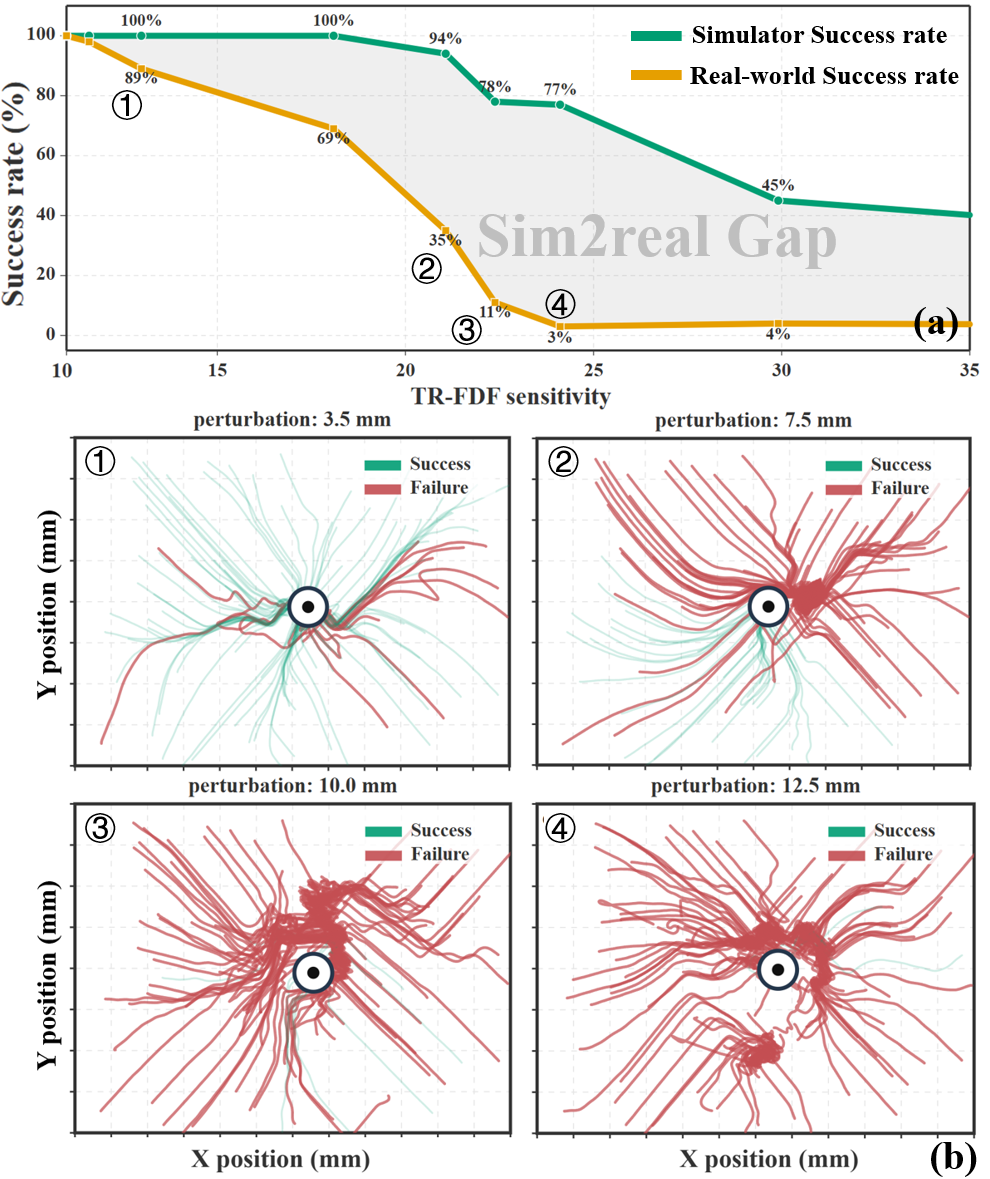}
\caption{Relationship between TR-FDF sensitivity and sim-to-real success under real-image lookup intervention. (a) Simulator and real-robot success rates as a function of TR-FDF sensitivity. (b) Real-robot XY trajectories under different perturbation strengths; as the perturbation increases, the trajectories gradually shift from concentrated convergence to divergent failure.}
\label{fig:b1}
\end{figure}

Fig.~\ref{fig:b1}(b) shows the corresponding change in trajectory behavior, from concentrated convergence to widespread off-target divergence. Under larger perturbations, the convergence points even separated into two groups. This result confirms our conclusion that TR-FDF sensitivity affects convergence multiplicatively rather than additively. This phenomenon becomes more evident in Section~V-D.2. Across all conditions, TR-FDF sensitivity showed a strong negative correlation with real-world success, approximately \(-0.92\).

We use a two-dimensional lookup table because extending the lookup to 6DoF would require an impractically large dataset. Even sampling only 50 positions along each degree of freedom would require approximately \(50^6\) real images, which is impractical to collect in a laboratory setting. This experiment therefore provides a low-confounding controlled test of the motion-dependent mismatch characterized by \(L_{\mathrm{TR}}\). Section~V-D.2 further provides a 6DoF perturbation test in the simulated environment.

\begin{table*}[!t]
\caption{Results of the real-image lookup intervention.}
\label{tab:expB1_lookup}
\centering
\resizebox{\textwidth}{!}{%
\begin{tabular}{ccccccc}
\toprule
Position error (mm) & FID & TR-FDF sensitivity  & Sim. success (98\% CI) & Real success (98\% CI) & Train-real gap & Terminal err. (mm) \\
\midrule
0 (clean) & 0.815 & 10.99 & 100/100 [94.9, 100.0] & 100/100 [94.9, 100.0] & 0.0 pp & \(1.86\pm0.11\) \\
2.5 & 1.100 & 11.59 & 100/100 [94.9, 100.0] & 98/100 [91.5, 99.6] & +2.0 pp & \(2.49\pm1.13\) \\
3.5 & 0.665 & 12.98 & 100/100 [94.9, 100.0] & 89/100 [79.6, 94.4] & +11.0 pp & \(2.27\pm1.77\) \\
5.0 & 0.665 & 18.09 & 100/100 [94.9, 100.0] & 69/100 [57.5, 78.5] & +31.0 pp & \(4.15\pm3.78\) \\
7.5 & 0.712 & 21.07 & 94/100 [85.9, 97.6] & 35/100 [24.9, 46.6] & +59.0 pp & \(7.92\pm5.15\) \\
10.0 & 0.732 & 22.38 & 78/100 [67.1, 86.1] & 11/100 [5.6, 20.4] & +67.0 pp & \(12.33\pm5.04\) \\
12.5 & 0.685 & 24.11 & 77/100 [66.0, 85.2] & 3/100 [0.9, 10.0] & +74.0 pp & \(12.77\pm5.69\) \\
15.0 & 0.856 & 29.91 & 45/100 [34.0, 56.5] & 4/100 [1.3, 11.4] & +41.0 pp & \(13.62\pm5.38\) \\
20.0 & 1.020 & 49.95 & 26/100 [17.2, 37.2] & 3/100 [0.9, 10.0] & +23.0 pp & \(17.12\pm6.18\) \\
25.0 & 0.628 & 54.82 & 16/100 [9.3, 26.2] & 4/100 [1.3, 11.4] & +12.0 pp & \(25.81\pm5.91\) \\
\bottomrule
\end{tabular}}
\end{table*}

\subsubsection{Learned-Simulator Structural Intervention}
The learned-simulator structural intervention examined whether the failure mode observed in Section~V-D.1 also arose in the complete learned-simulator and 6DoF deployment pipeline. The intervention preserved visible anatomy while altering how neighboring structures evolved with pose and maintaining comparable single-frame fidelity. Each condition was characterized by an approximate mean morphology perturbation \(m\).

The experiment used the same dual frozen-encoder evaluation protocol as Section~V-D.1, with real ultrasound images providing the reference feature-transition distributions. As shown in Table~\ref{tab:expB2_learned_sim}, the diagnostic metrics exhibited a clear system-level separation. Perturbed FID remained within 30.35--31.29, close to the unperturbed value of 29.66. Simulator success over the final 100 episodes also remained at 99/100--100/100. Meanwhile, TR-FDF sensitivity increased from 31.61 to 302.64, whereas real-world success decreased from 98/100 to 0/100.

The results were consistent with Section~V-D.1. Under comparable image quality, TR-FDF sensitivity was negatively associated with real-world success, and this relationship could not be explained by FID. The most informative interval was \(m\approx4.5\)--\(5.5\,\mathrm{mm}\). Within this interval, TR-FDF sensitivity increased from 133.31 to 262.64, while real-world success decreased from 100/100 to 2/100 despite near-perfect simulator success. This pattern also explains why some policies achieve high success rates in simulation yet remain difficult to transfer to real robots: the discrepancy reflects the influence of TR-FDF sensitivity.

At \(m\approx5.0\,\mathrm{mm}\), some trajectories converged, whereas others formed a separate divergent cluster. At \(m\approx5.5\,\mathrm{mm}\), TR-FDF sensitivity reached 262.64, real-world success decreased to 2/100, and the system became further destabilized.

Taken together, the two experiments in Section~V-D support three conclusions:
\begin{enumerate}
    \item Under the controlled interventions, higher TR-FDF sensitivity was associated with lower sim-to-real success, and this relationship could not be explained by FID. TR-FDF sensitivity therefore provides a diagnostic complementary to single-frame image quality.
    \item Higher TR-FDF sensitivity was associated with a large gap between simulator and real-robot success, explaining why some policies achieve high success rates in simulation yet remain difficult to transfer to real robots.
    \item The observed transition from convergence to separated divergent trajectories was consistent with the multiplicative effect of TR-FDF sensitivity on closed-loop convergence predicted in Section~III.
\end{enumerate}

\begin{table*}[!t]
\caption{Results of the learned-simulator structural intervention.}
\label{tab:expB2_learned_sim}
\centering
\resizebox{\textwidth}{!}{%
\begin{tabular}{ccccccc}
\toprule
Morphology error (mm) & FID & TR-FDF sensitivity  & Sim. success (98\% CI) & Real success (98\% CI) & Train-real gap & Terminal err. (mm/deg) \\
\midrule
0 (clean) & 29.66 & 31.61 & 100/100 [94.9, 100.0] & 98/100 [91.5, 99.6] & +2.0 pp & \(2.83\pm0.95\) / \(3.88\pm0.86^\circ\) \\
\(\approx2.0\) & 31.29 & 59.94 & 100/100 [94.9, 100.0] & 100/100 [94.9, 100.0] & 0.0 pp & \(2.63\pm0.61\) / \(3.76\pm1.04^\circ\) \\
\(\approx3.0\) & 31.23 & 123.49 & 100/100 [94.9, 100.0] & 97/100 [90.0, 99.1] & +3.0 pp & \(2.92\pm1.11\) / \(3.96\pm0.97^\circ\) \\
\(\approx4.5\) & 30.78 & 133.31 & 99/100 [93.1, 99.9] & 100/100 [94.9, 100.0] & -1.0 pp & \(2.78\pm0.25\) / \(3.77\pm0.89^\circ\) \\
\(\approx5.0\) & 31.15 & 183.75 & 100/100 [94.9, 100.0] & 73/100 [61.7, 81.9] & +27.0 pp & \(19.85\pm31.95\) / \(3.64\pm2.14^\circ\) \\
\(\approx5.5\) & 30.70 & 262.64 & 100/100 [94.9, 100.0] & 2/100 [0.4, 8.5] & +98.0 pp & \(48.13\pm20.69\) / \(5.17\pm1.84^\circ\) \\
\(\approx7.5\) & 30.35 & 302.64 & 100/100 [94.9, 100.0] & 0/100 [0.0, 5.1] & +100.0 pp & \(49.03\pm18.60\) / \(5.30\pm1.65^\circ\) \\
\bottomrule
\end{tabular}}
\end{table*}

\begin{figure}[!t]
\centering
\includegraphics[width=\columnwidth]{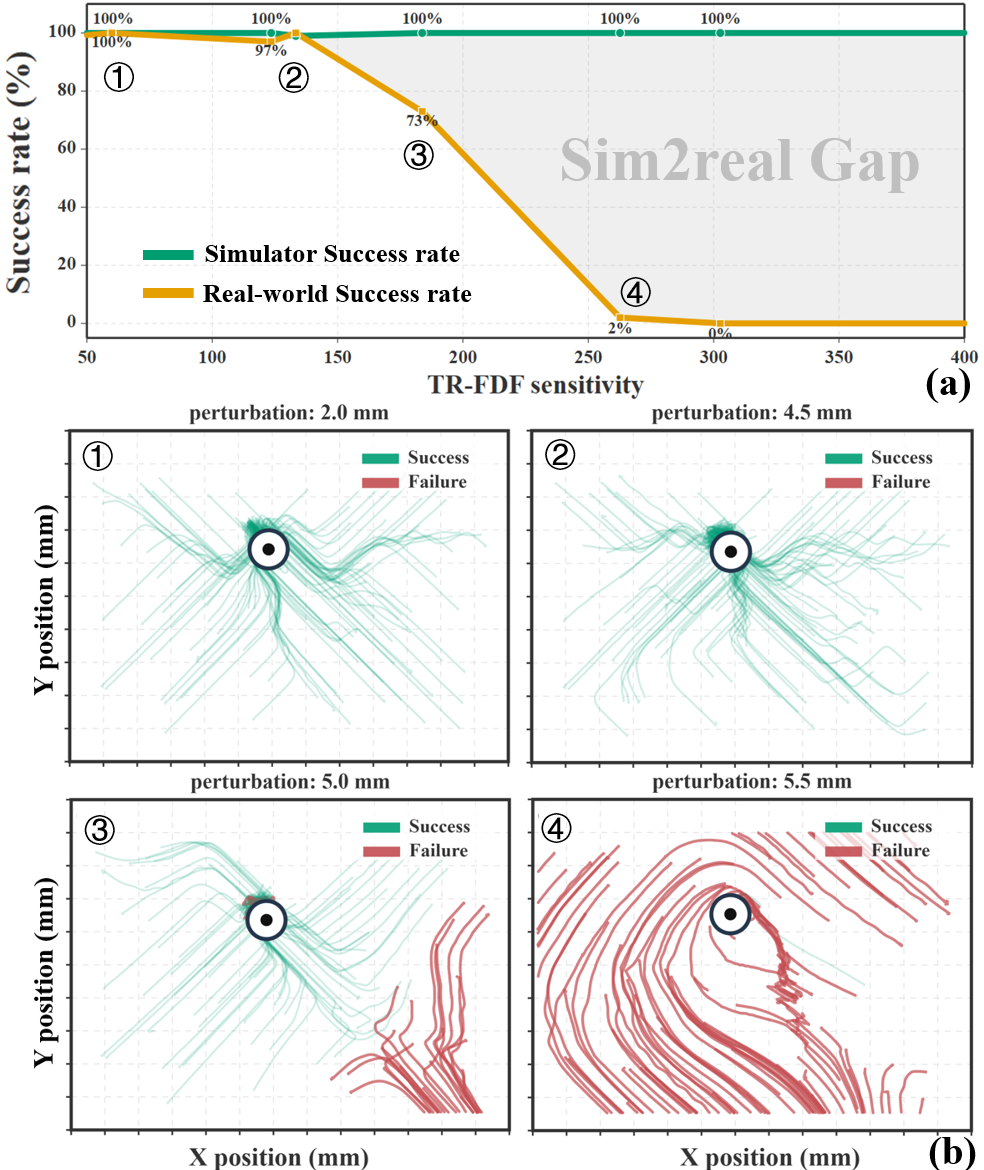}
\caption{Relationship between TR-FDF sensitivity and sim-to-real success under structural intervention in the learned simulator. (a) Simulator and real-robot success rates as a function of TR-FDF sensitivity. (b) Real-robot XY trajectories under different perturbation strengths; under large perturbations, convergent trajectories and a separate divergent cluster appear simultaneously. Green and red denote successful and failed episodes, respectively.}
\label{fig:b2}
\end{figure}

\subsection{Ablation Studies}
This section explains why the proposed observation generator is effective. The internal ablations examine how its design choices support the three criteria in Section~\ref{sec:criteria}: TR-FDF sensitivity, single-frame realism, and rollout efficiency. These experiments aim to identify the specific role of each module within the overall simulator design.

The ablation study has two levels. First, the component-level study in Section~\ref{sec:component_ablation} removes or weakens five major components of the full generator: fixed-\(\xfixed\) trajectory rendering, the shared structural intermediate domain, Stage-I shaping, the CFM generator, and reflow distillation. Second, the loss-level study in Section~\ref{sec:stage1_loss_ablation} removes each of the six Stage-I loss terms described in Section~IV-B to analyze their different roles in image generation.

\subsubsection{Component-Level Ablations with Real-Robot Deployment}
\label{sec:component_ablation}
The component-level study ablated five components of the observation generator to determine how they support TR-FDF sensitivity, single-frame realism, and rollout efficiency. Table~\ref{tab:expC_component_ablation} compares the full generator with five ablation variants. These variants use frame-level random \(\xfixed\), direct mask-to-US rendering without the shared structural domain, weakened Stage-I shaping, a GAN-based generator in place of the CFM generator, or no reflow distillation.

Each condition reports TR-FDF sensitivity, FID, real-robot success over 100 episodes, terminal errors, simulator-side success, and inference speed. Real-robot success is the final evaluation outcome, whereas TR-FDF sensitivity, FID, and inference speed diagnose dynamics degradation, loss of single-frame realism, and rollout efficiency, respectively.

\begin{table*}[!t]
\caption{Component-level ablation results for the proposed simulator.}
\label{tab:expC_component_ablation}
\centering
\resizebox{\textwidth}{!}{%
\begin{tabular}{llccccccc}
\toprule
Experiment & Ablation setting & TR-FDF sensitivity  & FID & Real success & Trans. err. (mm) & Rot. err. (deg) & Sim. success & Speed (Hz) \\
\midrule
Baseline & No ablation & 31.61 & 29.66 & 98/100 & \(2.83\pm0.95\) & \(3.88\pm0.86\) & 100\% & 67.1 \\
Ablation 1 & Random frame-level \(\xfixed\) & 50.78 & 43.36 & 55/100 & \(3.29\pm4.82\) & \(6.16\pm2.04\) & 100\% & 68.4 \\
Ablation 2 & No Stage-I shaping & 244.26 & 31.14 & 5/100 & \(20.44\pm17.62\) & \(11.98\pm5.80\) & 100\% & 76.4 \\
Ablation 3 & Weak Stage-I shaping & 164.61 & 53.05 & 17/100 & \(6.01\pm3.54\) & \(6.61\pm1.64\) & 100\% & 64.9 \\
Ablation 4 & GAN-based generator & 92.03 & 210.83 & 18/100 & \(5.95\pm7.79\) & \(10.93\pm4.73\) & 100\% & 130.7 \\
Ablation 5 & No reflow distillation & 30.43 & 27.83 & 99/100 & \(2.54\pm0.68\) & \(3.41\pm0.81\) & 100\% & 3.07 \\
\bottomrule
\end{tabular}}
\end{table*}

\begin{figure}[!t]
\centering
\includegraphics[width=\columnwidth]{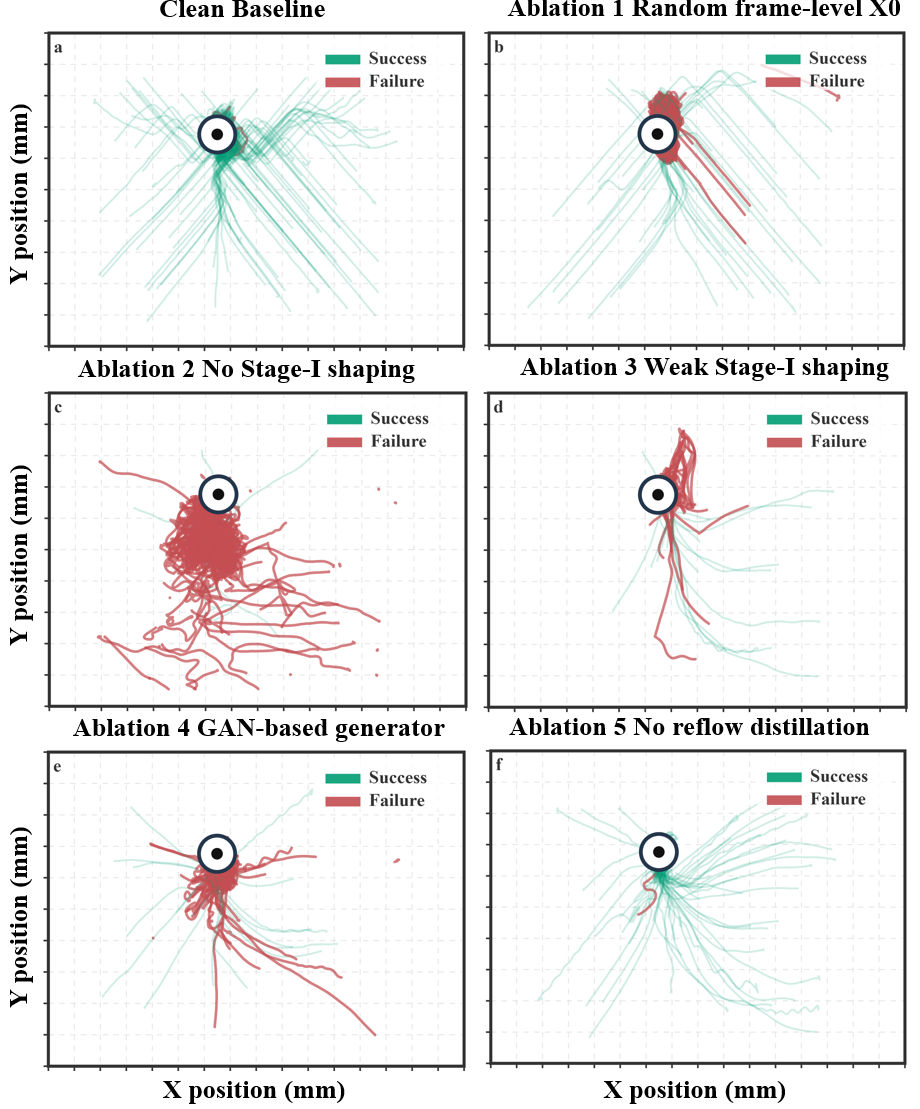}
\caption{Real-robot XY trajectories for the component-level ablations. Green and red denote successful and failed episodes, respectively; all panels use the same target and coordinate ranges.}
\label{fig:c1_trajectories}
\end{figure}

With frame-level random \(\xfixed\), real-robot success decreased from the baseline of 98/100 to 55/100. TR-FDF sensitivity increased from 31.61 to 50.78, FID increased from 29.66 to 43.36, and inference speed remained nearly unchanged. Figure~\ref{fig:c1_trajectories} shows the corresponding transition from concentrated convergence to more dispersed terminal states and off-target trajectories. These results indicate that fixing \(\xfixed\) along a trajectory helps maintain consistency between consecutive observations.

Ablation 2 removed the shared structural domain, increasing TR-FDF sensitivity to 244.26 and reducing real-robot success to 5/100. Its FID remained 31.14, close to the baseline value of 29.66. Thus, even with comparable single-frame quality, removing the shared structural domain severely disrupted feature dynamics and sim-to-real transfer. Ablation 3 weakened Stage-I shaping, increasing TR-FDF sensitivity to 164.61 and FID to 53.05 while reducing real-robot success to 17/100. These results indicate that the shared structural domain and Stage-I shaping are both important for maintaining pose-driven structural dynamics, cross-domain alignment, and downstream image generation.

Ablation 4 replaced the CFM generator with a GAN-based generator. Although inference speed increased from 67.1 to 130.7~Hz, FID degraded from 29.66 to 210.83, and real-robot success decreased to 18/100. This result demonstrates the importance of the CFM generator for maintaining single-frame realism and real-robot transfer performance.

After reflow distillation was completely removed in Ablation 5, TR-FDF sensitivity was 30.43 and FID was 27.83. Both metrics were comparable to or better than those of the full model, and real-robot success remained 99/100. However, inference speed decreased from 67.1 to 3.07~Hz, which was insufficient for efficient closed-loop policy rollouts. Reflow distillation therefore primarily improves observation-generation speed rather than TR-FDF sensitivity or single-frame realism.

All five ablations achieved 100\% best-100 simulator success, yet their real-world performance and rollout efficiency differed substantially. Overall, simulator-side learnability or any single diagnostic metric was insufficient to determine transferability. TR-FDF sensitivity, single-frame realism, and inference efficiency must therefore be evaluated jointly.

\subsubsection{Stage-I Loss-Level Ablations}
\label{sec:stage1_loss_ablation}
The component-level results showed that Stage-I shaping as a whole was necessary for real-robot transfer. The loss-level study further removed each of the six Stage-I loss terms to analyze how each constraint affects the observation model. We report FID and TR-FDF sensitivity for the final generated images to assess single-frame realism and task-relevant feature dynamics, respectively.

Table~\ref{tab:expC2_stage1_loss} presents the results under the shared evaluation protocol. Following Section~IV-B.3, the six losses are divided into two functional groups. The first preserves TR-FDF-related latent dynamics and includes \(\mathcal{L}_{\mathrm{dist}}\), \(\mathcal{L}_{\mathrm{smooth}}\), and \(\mathcal{L}_{\mathrm{invar}}\). The second prevents degenerate representations and establishes the shared structural space through \(\mathcal{L}_{\mathrm{seg}}\), \(\mathcal{L}_{\mathrm{align}}\), and \(\mathcal{L}_{\mathrm{reg}}\).

\begin{table}[!t]
\caption{Ablation results for the key Stage-I losses.}
\label{tab:expC2_stage1_loss}
\centering
\resizebox{\columnwidth}{!}{%
\begin{tabular}{lcc}
\toprule
Ablation setting & TR-FDF sensitivity  & FID \\
\midrule
Full Stage I & \textbf{31.61} & \textbf{29.66} \\
\multicolumn{3}{l}{\textit{TR-FDF-related latent dynamics}} \\
No distance matching & 35.01 & 29.98 \\
No mask-latent smoothness & 138.34 & 59.45 \\
No appearance invariance & 49.18 & 40.97 \\
\midrule
\multicolumn{3}{l}{\textit{Non-collapse and shared structural representation}} \\
No segmentation & 46.88 & 63.36 \\
No latent alignment & 35.43 & 39.12 \\
No latent regularization & 49.96 & 53.63 \\
\bottomrule
\end{tabular}}
\end{table}

For the first group, removing \(\mathcal{L}_{\mathrm{dist}}\) left FID nearly unchanged but increased TR-FDF sensitivity from 31.61 to 35.01. This indicates that the loss primarily constrains the relative latent geometry induced by motion. Removing \(\mathcal{L}_{\mathrm{smooth}}\) increased TR-FDF sensitivity to 138.34, the largest change among the six ablations, showing that local continuity is critical for stable feature dynamics. Removing \(\mathcal{L}_{\mathrm{invar}}\) increased TR-FDF sensitivity and FID to 49.18 and 40.97, respectively, indicating that appearance invariance helps suppress pose-independent latent variation.

For the second group, removing \(\mathcal{L}_{\mathrm{seg}}\), \(\mathcal{L}_{\mathrm{align}}\), and \(\mathcal{L}_{\mathrm{reg}}\) increased TR-FDF sensitivity to 46.88, 35.43, and 49.96, respectively. The corresponding FID values increased to 63.36, 39.12, and 53.63. These results support the respective roles of structural supervision, cross-modal latent alignment, and latent-scale regularization in establishing a non-degenerate shared structural representation.

This loss-level study uses offline metrics for mechanism analysis and does not perform complete real-robot evaluations for every variant. The system-level role of Stage I was assessed through the real-robot deployment in Section~\ref{sec:component_ablation}. Retraining the downstream policy after each loss ablation would introduce additional variation from policy optimization and robot execution, making direct attribution to an individual loss more difficult. Accordingly, this study analyzes how each loss affects observation-model properties without making separate claims about its contribution to real-robot success.

\subsection{Comparison with External Baselines}

This section examines whether the representative unpaired image-to-image (I2I) and video-translation methods introduced in Section~II-B can achieve low TR-FDF sensitivity and support zero-shot sim-to-real transfer of raw B-mode policies. The comparison evaluates whether the proposed architecture can preserve single-frame realism while jointly supporting task-relevant feature dynamics, rollout efficiency, simulator-side policy learning, and real-robot transfer.

\begin{table*}[!t]
\centering

\caption{Comparison with classical methods.}
\label{tab:expD_external_baselines}

\resizebox{\textwidth}{!}{%
\begin{tabular}{llcc c c c c}
\toprule
Screening stage & Method & FPS & FID & TR-FDF sensitivity & Sim. success & Real success & Final error (mm/deg) \\
\midrule
I2I-training failure & Recycle-GAN (2018)~\cite{Ban18} & 75.3 & 153.59 & -- & -- & -- & -- \\
I2I-training failure & Unsup-Recycle-GAN (2023)~\cite{Rec23} & 71.3 & 312.23 & -- & -- & -- & -- \\
\midrule
SAC-training failure & UNIT-DDPM (2021)~\cite{Sas21} & 10.6 & 34.87 & 47.23 & 0/100 & -- & -- \\
SAC-training failure & UNest (2024)~\cite{Pha24u} & 179.8 & 51.90 & 126.78 & 0/100 & -- & -- \\
SAC-training failure & STABLE (2025)~\cite{Sta25} & 105.5 & 45.31 & 93.71 & 0/100 & -- & -- \\
\midrule
Real-robot evaluation & CycleGAN (2017)~\cite{Zhu17} & 146.8 & 56.09 & 89.45 & 100/100 & 1/100 & $24.52\pm16.05$ / $17.66\pm4.34$ \\
Real-robot evaluation & CUT (2020)~\cite{Par20} & 130.6 & 62.97 & 42.61 & 100/100 & 7/100 & $19.40\pm13.31$ / $14.05\pm5.89$ \\
Real-robot evaluation & UVCGAN2 (2023)~\cite{Tor23} & 134.1 & \textbf{21.81} & 904.16 & 92/100 & 0/100 & $47.26\pm17.98$ / $16.06\pm4.90$ \\
Real-robot evaluation & UNSB (2024)~\cite{Kim24u} & 56.2 & 62.06 & 79.03 & 100/100 & 1/100 & $35.12\pm18.90$ / $17.12\pm4.87$ \\
\midrule
Real-robot evaluation & \textbf{Ours} & 67.1 & 29.66 & \textbf{31.61} & 100/100 & \textbf{98/100} & \textbf{$2.83\pm0.95$ / $3.88\pm0.86$} \\
\bottomrule
\end{tabular}%
}

\vspace{0.8em}

\includegraphics[width=\textwidth]{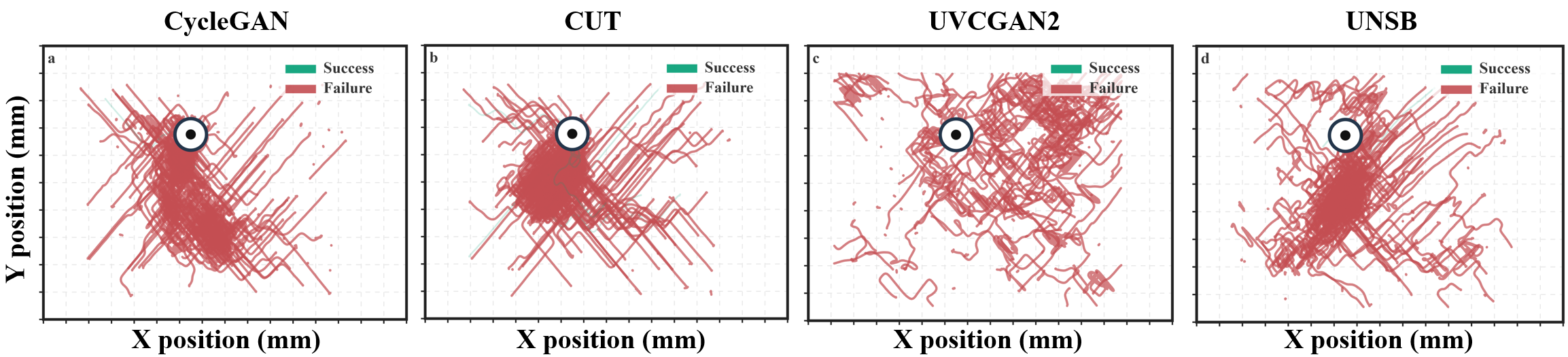}
\captionof{figure}{XY trajectories of external-baseline methods entering real-robot evaluation. Green and red denote successful and failed episodes, respectively, and the center marker indicates the common target. External baselines mainly produce off-target failure trajectories, whereas the trajectories of the proposed method shown in Fig.~\ref{fig:expA-traj} converge around the target.}
\label{fig:expD_trajectories}

\end{table*}

We compare representative GAN-based, Transformer-based, diffusion-based, and recurrent unpaired video-translation methods. A practical observation generator must pass three sequential screening levels. Level 1 requires the model to train stably and generate usable B-mode observations. Level 2 requires the generated observation sequences to support stable policy learning in the simulator. Level 3 requires the resulting policy to retain closed-loop convergence on the real robot. If a method fails at an earlier level, it is not evaluated at subsequent levels. The ``Screening stage'' column in Table~\ref{tab:expD_external_baselines} indicates the level at which each method failed or entered evaluation.

Table~\ref{tab:expD_external_baselines} and Fig.~\ref{fig:expD_trajectories} show that determining whether an observation generator is suitable for zero-shot sim-to-real transfer is a sequential screening problem that cannot be resolved using a single generation metric. First, I2I model convergence did not imply policy learnability. Although UNIT-DDPM, UNest, and STABLE generated images, none supported SAC convergence. Second, simulator-side policy success did not imply transfer to the real environment. CycleGAN, CUT, and UNSB all achieved 100/100 SAC success, but their real-robot success rates were only 1--7/100.

UVCGAN2 provides the most representative counterexample. It achieved the best FID in the table (21.81) and a simulator-side success rate of 92/100, but its TR-FDF sensitivity was the highest at 904.16, and none of its real-robot trials succeeded. These results show that neither single-frame distributional similarity nor simulator-side policy convergence can replace the evaluation of task-relevant feature dynamics. In contrast, CUT achieved the lowest TR-FDF sensitivity among the external baselines (42.61), but its FID was 62.97 and its real-robot success rate was only 7/100. UNIT-DDPM also achieved a relatively low TR-FDF sensitivity of 47.23 but failed during SAC training. These results show that FID and TR-FDF sensitivity diagnose different properties of the observation model.

Within this comparison, only the proposed simulator jointly achieved low TR-FDF sensitivity (31.61), high single-frame realism (FID of 29.66), usable rollout speed (67.1 FPS), stable SAC training (100/100), and reliable zero-shot real-robot deployment (98/100). The proposed architecture therefore maintained complete observation-generation and transfer capability across all three sequential screening levels.

\subsection{Task-Level Simulator Consistency under Disturbances}
This section tests whether a policy trained in the simulator retains the expected closed-loop behavior under corresponding real-world perturbations. It therefore evaluates whether the simulator can represent task-relevant observation changes induced by these perturbations. Section~V-G.1 evaluates contact-force variation, whereas Section~V-G.2 evaluates sustained phantom motion. Neither evaluation uses real-world fine-tuning or online adaptation.

\subsubsection{Deployment Consistency under Contact-Force Variation}
Changing contact force alters tissue compression and the resulting B-mode appearance, providing a direct test of perturbation consistency in the observation model. We evaluated real-robot success at 5~N and 15~N contact force, while holding all other training and evaluation settings fixed.

\begin{table}[!t]
\caption{Real-robot zero-shot deployment results under contact-force variation.}
\label{tab:expE_force}
\centering
\begin{tabular}{ccc}
\toprule
Contact force (N) & Real-robot success & Failure type \\
\midrule
5 & 94/100 & 6 timeouts \\
15 & 98/100 & 2 timeouts \\
\bottomrule
\end{tabular}
\end{table}

The policy achieved 94.0\% and 98.0\% success on the real robot at 5~N and 15~N, respectively. Within the tested 5--15~N range, the task-relevant observations provided by the proposed simulator supported zero-shot deployment under contact-force variation.

\subsubsection{High-Disturbance Tracking under Phantom Motion}
\begin{figure*}[!t]
\centering
\includegraphics[width=\textwidth]{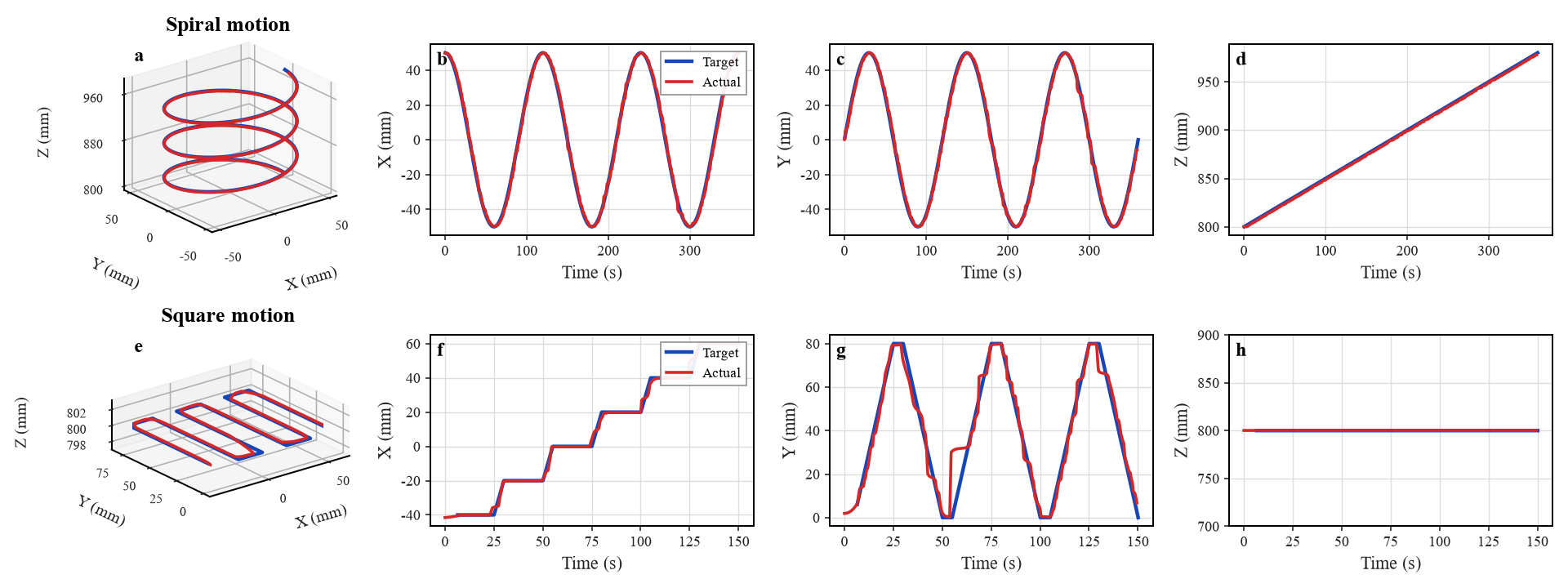}
\caption{Real-robot tracking results when the phantom moves along spiral and square trajectories. Blue denotes the target trajectory, and red denotes the actual probe trajectory.}
\label{fig:expE_motion}
\end{figure*}

We next fixed the simulator-trained policy and drove the phantom along predefined spiral and square trajectories, testing whether the robot could maintain closed-loop tracking under sustained target-pose variation. Figure~\ref{fig:expE_motion} compares the target and actual TCP trajectories in three dimensions and along their X/Y/Z components. The actual probe trajectory follows both target motions; notably, the policy continues to adjust through the piecewise direction changes of the square trajectory. These results show that the simulator-trained policy can operate in a real environment with substantial target-motion disturbances.

\section{Discussion and Limitations}
\subsection{Discussion}
This study shows that, for closed-loop ultrasound policies that take raw B-mode images as input, sim-to-real transfer depends not only on single-frame realism but also on whether probe motion induces consistent task-relevant feature changes in simulation and the real system. The analysis in Section~III shows that the local sensitivity of TR-FDF mismatch to probe motion consumes the contraction margin of the real closed loop, whereas static image bias and motion-independent perturbations primarily affect the residual error near the target. The two controlled interventions further showed that, with FID largely unchanged, increasing TR-FDF sensitivity was accompanied by lower real-robot success and a separation between simulator-side success and real-world deployment outcomes.

Motivated by this observation, we developed an ultrasound simulator for raw-B-mode policy training and achieved a 97.5\% success rate over 400 zero-shot real-robot deployments. The shared structural intermediate domain and Stage-I shaping preserve pose-driven structural changes; fixed-\(\xfixed\) rendering reduces trajectory-level sampling variation unrelated to probe motion; and CFM and reflow address B-mode appearance and rollout efficiency, respectively. Ablation studies and external-baseline comparisons indicate that an observation generator must provide task-relevant feature dynamics, single-frame realism, and inference speed, rather than being characterized by any single metric.

\subsection{Limitations}
TR-FDF sensitivity is intended to complement rather than replace single-frame image-quality measures. Errors in generated structure or appearance may affect both FID and task-feature transitions, so the two metrics may remain correlated. The controlled interventions showed that TR-FDF sensitivity captured transfer degradation not explained by FID, but did not establish statistical independence between the metrics. Accordingly, observation generators should be evaluated using TR-FDF sensitivity jointly with single-frame realism and rollout efficiency.

The local analysis in Section~III is, in principle, applicable to sensor modalities in which actions induce subsequent observation changes and the corresponding local boundedness conditions hold. However, the TR-FDF proxy, observation-generator architecture, and controlled validation in this study were developed for robotic ultrasound. Thus, the present work provides controlled intervention evidence in the ultrasound setting, but does not yet establish the empirical applicability of the framework to other observation modalities. Extending this analysis to other action-conditioned observation channels is an important direction for future work.

\balance
\bibliographystyle{IEEEtran}
\bibliography{TRO_references}

@article{Hua21,
  author  = {Huang, Yanwei and Xiao, Wei and Wang, Chuyang and Liu, Hengli and Huang, Rui-Peng and Sun, Zhenglong},
  title   = {Towards Fully Autonomous Ultrasound Scanning Robot With Imitation Learning Based on Clinical Protocols},
  journal = {IEEE Robotics and Automation Letters},
  year    = {2021},
  volume  = {6},
  number  = {2},
  pages   = {3671--3678},
  doi     = {10.1109/LRA.2021.3064283}
}

@article{Su24,
  author  = {Su, Kang and Liu, Jingwei and Ren, Xiaoqi and Huo, Yingxiang and Du, Guanglong and Zhao, Wei and Wang, Xueqian and Liang, Bin and Li, Di and Liu, Peter Xiaoping},
  title   = {A Fully Autonomous Robotic Ultrasound System for Thyroid Scanning},
  journal = {Nature Communications},
  year    = {2024},
  volume  = {15},
  number  = {1},
  note    = {Art. no. 4004},
  doi     = {10.1038/s41467-024-48421-y}
}

@article{Jia25,
  author  = {Jiang, Haojun and Zhao, Andrew and Yang, Qian and Yan, Xiangjie and Wang, Teng and Wang, Yulin and Jia, Ning and Wang, Jiangshan and Wu, Guokun and Yue, Yang and Luo, Shaqi and Wang, Huanqian and Ren, Ling and Chen, Siming and Liu, Pan and Yao, Guocai and Yang, Wenming and Song, Shiji and Li, Xiang and He, Kunlun and Huang, Gao},
  title   = {Towards Expert-Level Autonomous Carotid Ultrasonography with Large-Scale Learning-Based Robotic System},
  journal = {Nature Communications},
  year    = {2025},
  volume  = {16},
  number  = {1},
  note    = {Art. no. 7893},
  doi     = {10.1038/s41467-025-62865-w}
}

@inproceedings{Has20,
  author    = {Hase, H. and Azampour, Mohammad Farid and Tirindelli, M. and Paschali, Magdalini and Simson, Walter and Fatemizadeh, E. and Navab, Nassir},
  title     = {Ultrasound-Guided Robotic Navigation with Deep Reinforcement Learning},
  booktitle = {2020 IEEE/RSJ International Conference on Intelligent Robots and Systems},
  year      = {2020},
  pages     = {5534--5541},
  doi       = {10.1109/IROS45743.2020.9340913}
}

@inproceedings{Li21,
  author    = {Li, Keyu and Wang, Jian and Xu, Yangxin and Qin, Hao and Liu, Dongsheng and Liu, Li and Meng, M. Q.-H.},
  title     = {Autonomous Navigation of an Ultrasound Probe Towards Standard Scan Planes with Deep Reinforcement Learning},
  booktitle = {2021 IEEE International Conference on Robotics and Automation},
  year      = {2021},
  pages     = {8302--8308},
  doi       = {10.1109/ICRA48506.2021.9561295}
}

@article{Li21b,
  author  = {Li, Keyu and Xu, Yangxin and Wang, Jian and Ni, Dong and Liu, Li and Meng, M. Q.-H.},
  title   = {Image-Guided Navigation of a Robotic Ultrasound Probe for Autonomous Spinal Sonography Using a Shadow-Aware Dual-Agent Framework},
  journal = {IEEE Transactions on Medical Robotics and Bionics},
  year    = {2022},
  volume  = {4},
  number  = {1},
  pages   = {130--144},
  doi     = {10.1109/TMRB.2021.3127015}
}

@article{Bi22,
  author  = {Bi, Yuanwei and Jiang, Zhongliang and Gao, Yuan and Wendler, Thomas and Karlas, Angelos and Navab, Nassir},
  title   = {{VesNet-RL}: Simulation-Based Reinforcement Learning for Real-World {US} Probe Navigation},
  journal = {IEEE Robotics and Automation Letters},
  year    = {2022},
  volume  = {7},
  number  = {3},
  pages   = {6638--6645},
  doi     = {10.1109/LRA.2022.3176112}
}

@inproceedings{Hu24,
  author    = {Hu, Yicheng and Huang, Yixuan and Song, Anthony and Jones, Craig K. and Zbijewski, Wojciech and Siewerdsen, Jeffrey H. and Basar, Burcu and Helm, Patrick A. and Uneri, Ali},
  title     = {Probe Positioning for Robot-Assisted Intraoperative Ultrasound Imaging Using Deep Reinforcement Learning},
  booktitle = {Medical Imaging 2024: Image-Guided Procedures, Robotic Interventions, and Modeling},
  year      = {2024},
  volume    = {12928},
  publisher = {SPIE},
  paper     = {1292803},
  type      = {Art. no.},
  doi       = {10.1117/12.3006918}
}

@inproceedings{Tom21,
  author    = {Tomar, Devavrat and Zhang, Lin and Portenier, Tiziano and Goksel, Orcun},
  title     = {Content-Preserving Unpaired Translation from Simulated to Realistic Ultrasound Images},
  booktitle = {Medical Image Computing and Computer-Assisted Intervention -- MICCAI 2021},
  year      = {2021},
  pages     = {659--669},
  publisher = {Springer},
  doi       = {10.1007/978-3-030-87237-3_63}
}

@article{Vit20,
  author  = {Vitale, Santiago and Orlando, Jos{\'e} Ignacio and Iarussi, Emmanuel and Larrabide, Ignacio},
  title   = {Improving Realism in Patient-Specific Abdominal Ultrasound Simulation Using {CycleGANs}},
  journal = {International Journal of Computer Assisted Radiology and Surgery},
  year    = {2020},
  volume  = {15},
  number  = {1},
  pages   = {183--192},
  doi     = {10.1007/s11548-019-02046-5}
}

@article{Vit25,
  author  = {Vitale, Santiago and Orlando, Jos{\'e} Ignacio and Iarussi, Emmanuel and D{\'i}az, Alejandro and Larrabide, Ignacio},
  title   = {Improving Realism in Abdominal Ultrasound Simulation Combining a Segmentation-Guided Loss and Polar Coordinates Training},
  journal = {Medical Physics},
  year    = {2025},
  volume  = {52},
  number  = {9},
  pages   = {4540--4556},
  doi     = {10.1002/mp.17801}
}

@article{Jia23,
  author  = {Jiang, Zhongliang and Bi, Yuanwei and Zhou, Mingchuan and Hu, Ying and Burke, Michael and Navab, Nassir},
  title   = {Intelligent Robotic Sonographer: Mutual Information-Based Disentangled Reward Learning from Few Demonstrations},
  journal = {The International Journal of Robotics Research},
  year    = {2024},
  volume  = {43},
  number  = {7},
  pages   = {981--1002},
  doi     = {10.1177/02783649231223547}
}

@article{Vel22,
  author  = {Velikova, Yordanka and Simson, Walter and Salehi, M. and Azampour, Mohammad Farid and Paprottka, Peter and Navab, Nassir},
  title   = {{CACTUSS}: Common Anatomical {CT-US} Space for {US} Examinations},
  journal = {International Journal of Computer Assisted Radiology and Surgery},
  year    = {2024},
  volume  = {19},
  number  = {5},
  pages   = {861--869},
  doi     = {10.1007/s11548-024-03060-y}
}

@inproceedings{Vel23,
  author       = {Velikova, Yordanka and Azampour, Mohammad Farid and Simson, Walter and Duque, Vanessa Gonzalez and Navab, Nassir},
  title        = {{LOTUS}: Learning to Optimize Task-Based {US} Representations},
  booktitle    = {Medical Image Computing and Computer-Assisted Intervention -- MICCAI 2023},
  year         = {2023},
  pages        = {435--445},
  publisher    = {Springer},
  doi          = {10.48550/arXiv.2307.16021},
  eprint       = {2307.16021},
  archivePrefix = {arXiv}
}

@misc{Yan20,
  author        = {Yan, Mengyuan and Sun, Qingyun and Frosio, I. and Tyree, Stephen and Kautz, Jan},
  title         = {How to Close Sim-Real Gap? Transfer with Segmentation!},
  year          = {2020},
  howpublished  = {arXiv preprint arXiv:2005.07695},
  eprint        = {2005.07695},
  archivePrefix = {arXiv},
  url           = {https://arxiv.org/abs/2005.07695}
}

@inproceedings{Jam18,
  author    = {James, Stephen and Wohlhart, Paul and Kalakrishnan, Mrinal and Kalashnikov, Dmitry and Irpan, Alex and Ibarz, Julian and Levine, Sergey and Hadsell, Raia and Bousmalis, Konstantinos},
  title     = {Sim-To-Real via Sim-To-Sim: Data-Efficient Robotic Grasping via Randomized-To-Canonical Adaptation Networks},
  booktitle = {2019 IEEE/CVF Conference on Computer Vision and Pattern Recognition},
  year      = {2019},
  pages     = {12619--12629},
  doi       = {10.1109/CVPR.2019.01291}
}

@article{Mur21,
  author  = {Muratore, Fabio and Ramos, Fabio and Turk, Greg and Yu, Wenhao and Gienger, Michael and Peters, Jan},
  title   = {Robot Learning from Randomized Simulations: A Review},
  journal = {Frontiers in Robotics and AI},
  year    = {2022},
  volume  = {9},
  note    = {Art. no. 799893},
  doi     = {10.3389/frobt.2022.799893}
}

@article{Zha22h,
  author  = {Zhang, Xiaoshuai and Chen, Rui and Li, Ang and Xiang, Fanbo and Qin, Yuzhe and Gu, Jiayuan and Ling, Zhan and Liu, Minghua and Zeng, Peiyu and Han, Songfang and Huang, Zhiao and Mu, Tongzhou and Xu, Jing and Su, Hao},
  title   = {Close the Optical Sensing Domain Gap by Physics-Grounded Active Stereo Sensor Simulation},
  journal = {IEEE Transactions on Robotics},
  year    = {2023},
  volume  = {39},
  number  = {4},
  pages   = {2429--2447},
  doi     = {10.1109/TRO.2023.3235591}
}

@inproceedings{Nar21,
  author    = {Narang, Yashraj S. and Sundaralingam, Balakumar and Macklin, Miles and Mousavian, Arsalan and Fox, Dieter},
  title     = {Sim-to-Real for Robotic Tactile Sensing via Physics-Based Simulation and Learned Latent Projections},
  booktitle = {2021 IEEE International Conference on Robotics and Automation},
  year      = {2021},
  pages     = {6444--6451},
  doi       = {10.1109/ICRA48506.2021.9561969}
}

@inproceedings{Jen04,
  author    = {Jensen, J. A.},
  title     = {Simulation of Advanced Ultrasound Systems Using {Field II}},
  booktitle = {Proceedings of the 2nd IEEE International Symposium on Biomedical Imaging: Nano to Macro},
  year      = {2004},
  volume    = {1},
  pages     = {636--639},
  doi       = {10.1109/ISBI.2004.1398618}
}

@inproceedings{Due25,
  author    = {Duelmer, Felix and Azampour, Mohammad Farid and Wysocki, Magdalena and Navab, Nassir},
  title     = {{UltraRay}: Introducing Full-Path Ray Tracing in Physics-Based Ultrasound Simulation},
  booktitle = {Medical Image Computing and Computer-Assisted Intervention -- MICCAI 2025},
  year      = {2025},
  pages     = {653--662},
  publisher = {Springer},
  doi       = {10.1007/978-3-032-04937-7_62}
}

@article{Sol25,
  author  = {Solano-Cordero, Cindy M. and Encina-Baranda, N. and Perez-Liva, M. and Herraiz, Joaquin L.},
  title   = {Recent Advances in {B-Mode} Ultrasound Simulators},
  journal = {Applied Sciences},
  year    = {2025},
  volume  = {15},
  number  = {23},
  note    = {Art. no. 12535},
  doi     = {10.3390/app152312535}
}

@inproceedings{Sha08,
  author    = {Shams, R. and Hartley, R. and Navab, Nassir},
  title     = {Real-Time Simulation of Medical Ultrasound from {CT} Images},
  booktitle = {Medical Image Computing and Computer-Assisted Intervention -- MICCAI 2008},
  year      = {2008},
  pages     = {734--741},
  publisher = {Springer},
  doi       = {10.1007/978-3-540-85990-1_88}
}

@inproceedings{Zhu17,
  author    = {Zhu, Jun-Yan and Park, Taesung and Isola, Phillip and Efros, Alexei A.},
  title     = {Unpaired Image-to-Image Translation Using Cycle-Consistent Adversarial Networks},
  booktitle = {2017 IEEE International Conference on Computer Vision},
  year      = {2017},
  pages     = {2242--2251},
  doi       = {10.1109/ICCV.2017.244}
}

@inproceedings{Fre25,
  author    = {Freiche, Benoit and El-Khoury, Anthony and Nasiri-Sarvi, Ali and Hosseini, Mahdi S. and Garcia, Damien and Basarab, Adrian and Boily, Mathieu and Rivaz, Hassan},
  title     = {Ultrasound Image Generation Using Latent Diffusion Models},
  booktitle = {Medical Imaging 2025: Ultrasonic Imaging and Tomography},
  year      = {2025},
  volume    = {13412},
  paper     = {134121F},
  type      = {Art. no.},
  editor    = {Boehm, Christian and Mehrmohammadi, Mohammad},
  publisher = {SPIE},
  doi       = {10.1117/12.3047788}
}

@inproceedings{Dom24,
  author    = {Dominguez, Marina and Velikova, Yordanka and Navab, Nassir and Azampour, Mohammad Farid},
  title     = {Diffusion as Sound Propagation: Physics-Inspired Model for Ultrasound Image Generation},
  booktitle = {Medical Image Computing and Computer-Assisted Intervention -- MICCAI 2024},
  year      = {2024},
  pages     = {613--623},
  publisher = {Springer},
  doi       = {10.1007/978-3-031-72083-3_57}
}

@inproceedings{Jud25,
  author    = {Judge, Thierry and Duchateau, Nicolas and Faraz, Khuram and Jodoin, Pierre-Marc and Bernard, Olivier},
  title     = {Generation of Realistic Cardiac Ultrasound Sequences with Ground Truth Motion and Speckle Decorrelation},
  booktitle = {2025 IEEE International Ultrasonics Symposium},
  year      = {2025},
  pages     = {1--4},
  doi       = {10.1109/IUS62464.2025.11201671}
}

@inproceedings{Wan18,
  author    = {Wang, Ting-Chun and Liu, Ming-Yu and Zhu, Jun-Yan and Liu, Guilin and Tao, Andrew and Kautz, Jan and Catanzaro, Bryan},
  title     = {Video-to-Video Synthesis},
  booktitle = {Advances in Neural Information Processing Systems},
  year      = {2018},
  volume    = {31},
  pages     = {1152--1164}
}

@inproceedings{Ban18,
  author    = {Bansal, Aayush and Ma, Shugao and Ramanan, Deva and Sheikh, Yaser},
  title     = {{Recycle-GAN}: Unsupervised Video Retargeting},
  booktitle = {Computer Vision -- ECCV 2018},
  year      = {2018},
  volume    = {11209},
  pages     = {122--138},
  publisher = {Springer},
  doi       = {10.1007/978-3-030-01228-1_8}
}

@inproceedings{Rec23,
  author    = {Wang, Kaihong and Akash, Kumar and Misu, Teruhisa},
  title     = {Learning Temporally and Semantically Consistent Unpaired Video-to-Video Translation Through Pseudo-Supervision from Synthetic Optical Flow},
  booktitle = {Proceedings of the AAAI Conference on Artificial Intelligence},
  year      = {2022},
  volume    = {36},
  number    = {3},
  pages     = {2477--2486},
  doi       = {10.1609/aaai.v36i3.20148}
}

@inproceedings{Par20,
  author    = {Park, Taesung and Efros, Alexei A. and Zhang, Richard and Zhu, Jun-Yan},
  title     = {Contrastive Learning for Unpaired Image-to-Image Translation},
  booktitle = {Computer Vision -- ECCV 2020},
  year      = {2020},
  volume    = {12354},
  pages     = {319--345},
  publisher = {Springer},
  doi       = {10.1007/978-3-030-58545-7_19}
}

@misc{Sas21,
  author        = {Sasaki, Hiroshi and Willcocks, Chris G. and Breckon, Toby P.},
  title         = {{UNIT-DDPM}: Unpaired Image Translation with Denoising Diffusion Probabilistic Models},
  year          = {2021},
  howpublished  = {arXiv preprint arXiv:2104.05358},
  eprint        = {2104.05358},
  archivePrefix = {arXiv},
  url           = {https://arxiv.org/abs/2104.05358}
}

@misc{Tor23,
  author        = {Torbunov, Dmitrii and Huang, Yi and Tseng, Huan-Hsin and Yu, Haiwang and Huang, Jin and Yoo, Shinjae and Lin, Meifeng and Viren, Brett and Ren, Yihui},
  title         = {{UVCGAN v2}: An Improved Cycle-Consistent {GAN} for Unpaired Image-to-Image Translation},
  year          = {2023},
  howpublished  = {arXiv preprint arXiv:2303.16280},
  eprint        = {2303.16280},
  archivePrefix = {arXiv},
  url           = {https://arxiv.org/abs/2303.16280}
}

@inproceedings{Kim24u,
  author    = {Kim, Beomsu and Kwon, Gihyun and Kim, Kwanyoung and Ye, Jong Chul},
  title     = {Unpaired Image-to-Image Translation via Neural {Schr{\"o}dinger} Bridge},
  booktitle = {International Conference on Learning Representations},
  year      = {2024},
  eprint    = {2305.15086},
  archivePrefix = {arXiv},
  url       = {https://openreview.net/forum?id=uQBW7ELXfO}
}

@inproceedings{Pha24u,
  author    = {Phan, Vu Minh Hieu and Xie, Yutong and Zhang, Bowen and Qi, Yuankai and Liao, Zhibin and Perperidis, Antonios and Phung, Son Lam and Verjans, Johan W. and To, Minh-Son},
  title     = {Structural Attention: Rethinking Transformer for Unpaired Medical Image Synthesis},
  booktitle = {Medical Image Computing and Computer-Assisted Intervention -- MICCAI 2024},
  year      = {2024},
  pages     = {690--700},
  publisher = {Springer},
  doi       = {10.1007/978-3-031-72104-5_66}
}

@article{Sta25,
  author  = {You, Joshua Yedam and Eom, Minho and Choi, Tae-Ik and Cho, Eun-Seo and Choi, Jieun and Lee, Minyoung and Shin, Changyeop and Moon, Jieun and Kim, Eunji and Kim, Pilhan and Kim, Cheol-Hee and Yoon, Young-Gyu},
  title   = {Preserving Spatial and Quantitative Information in Unpaired Biomedical Image-to-Image Translation},
  journal = {Cell Reports Methods},
  year    = {2025},
  volume  = {5},
  number  = {6},
  note    = {Art. no. 101074},
  doi     = {10.1016/j.crmeth.2025.101074}
}

@inproceedings{Par19,
  author    = {Park, Kwanyong and Woo, Sanghyun and Kim, Dahun and Cho, Donghyeon and Kweon, In-So},
  title     = {Preserving Semantic and Temporal Consistency for Unpaired Video-to-Video Translation},
  booktitle = {Proceedings of the 27th ACM International Conference on Multimedia},
  year      = {2019},
  pages     = {151--159},
  doi       = {10.1145/3343031.3350864}
}

@inproceedings{Riv21,
  author    = {Rivoir, Dominik and Pfeiffer, Micha and Docea, R. and Kolbinger, F. and Riediger, C. and Weitz, Juergen and Speidel, S.},
  title     = {Long-Term Temporally Consistent Unpaired Video Translation from Simulated Surgical {3D} Data},
  booktitle = {2021 IEEE/CVF International Conference on Computer Vision},
  year      = {2021},
  pages     = {3323--3333},
  doi       = {10.1109/ICCV48922.2021.00333}
}

@inproceedings{Liu24m,
  author    = {Liu, Jiahe and Qu, Youran and Yan, Qi and Zeng, Xiaohui and Wang, Lele and Liao, Renjie},
  title     = {Fr{\'e}chet Video Motion Distance: A Metric for Evaluating Motion Consistency in Videos},
  booktitle = {First Workshop on Controllable Video Generation at ICML 2024},
  year      = {2024},
  note      = {Workshop paper},
  eprint    = {2407.16124},
  archivePrefix = {arXiv},
  url       = {https://arxiv.org/abs/2407.16124}
}

@inproceedings{Tod12,
  author    = {Todorov, Emanuel and Erez, Tom and Tassa, Yuval},
  title     = {{MuJoCo}: A Physics Engine for Model-Based Control},
  booktitle = {2012 IEEE/RSJ International Conference on Intelligent Robots and Systems},
  year      = {2012},
  pages     = {5026--5033},
  doi       = {10.1109/IROS.2012.6386109}
}

@inproceedings{Aga20,
  author    = {Agarwal, Arpit and Man, Timothy and Yuan, Wenzhen},
  title     = {Simulation of Vision-Based Tactile Sensors Using Physics-Based Rendering},
  booktitle = {2021 IEEE International Conference on Robotics and Automation},
  year      = {2021},
  pages     = {1--7},
  doi       = {10.1109/ICRA48506.2021.9561122}
}

@inproceedings{Ngo21,
  author    = {Ngo, Anthony and Bauer, Max Paul and Resch, Michael},
  title     = {A Multi-Layered Approach for Measuring the Simulation-to-Reality Gap of Radar Perception for Autonomous Driving},
  booktitle = {2021 IEEE International Intelligent Transportation Systems Conference},
  year      = {2021},
  pages     = {4008--4014},
  doi       = {10.1109/ITSC48978.2021.9564521}
}

@inproceedings{Dea19,
  author       = {Dean, Sarah and Matni, Nikolai and Recht, Benjamin and Ye, Vickie},
  title        = {Robust Guarantees for Perception-Based Control},
  booktitle    = {Proceedings of the 2nd Conference on Learning for Dynamics and Control},
  year         = {2020},
  volume       = {120},
  series       = {Proceedings of Machine Learning Research},
  pages        = {350--360},
  publisher    = {PMLR},
  url          = {https://proceedings.mlr.press/v120/dean20a.html}
}

@inproceedings{Dea20b,
  author       = {Dean, Sarah and Recht, Benjamin},
  title        = {Certainty Equivalent Perception-Based Control},
  booktitle    = {Proceedings of the 3rd Conference on Learning for Dynamics and Control},
  year         = {2021},
  volume       = {144},
  series       = {Proceedings of Machine Learning Research},
  pages        = {399--411},
  publisher    = {PMLR},
  url          = {https://proceedings.mlr.press/v144/dean21a.html}
}

@inproceedings{Cho22b,
  author       = {Chou, Glen and Ozay, Necmiye and Berenson, Dmitry},
  title        = {Safe Output Feedback Motion Planning from Images via Learned Perception Modules and Contraction Theory},
  booktitle    = {Workshop on the Algorithmic Foundations of Robotics XV},
  year         = {2022},
  pages        = {349--367},
  eprint       = {2206.06553},
  archivePrefix = {arXiv},
  doi          = {10.48550/arXiv.2206.06553},
  url          = {https://arxiv.org/abs/2206.06553}
}

@article{Dav21,
  author  = {Davydov, Alexander and Jafarpour, Saber and Bullo, Francesco},
  title   = {Non-Euclidean Contraction Theory for Robust Nonlinear Stability},
  journal = {IEEE Transactions on Automatic Control},
  year    = {2022},
  volume  = {67},
  number  = {12},
  pages   = {6667--6681},
  doi     = {10.1109/TAC.2022.3183966}
}

@article{Yuk10,
  author  = {Yuksel, Serdar and Linder, Tamas},
  title   = {Optimization and Convergence of Observation Channels in Stochastic Control},
  journal = {SIAM Journal on Control and Optimization},
  year    = {2012},
  volume  = {50},
  number  = {2},
  pages   = {864--887},
  doi     = {10.1137/100808976}
}

@misc{Dem25,
  author        = {Demirci, Yunus Emre and Kara, Ali Devran and Yuksel, Serdar},
  title         = {Sensitivity of Filter Kernels and Robustness Bounds to Transition and Measurement Kernel Perturbations in Partially Observable Stochastic Control},
  year          = {2025},
  howpublished  = {arXiv preprint arXiv:2508.10658},
  eprint        = {2508.10658},
  archivePrefix = {arXiv},
  url           = {https://arxiv.org/abs/2508.10658}
}

@misc{Kra26,
  author        = {Kraske, Benjamin and Ho, Qi Heng and Rossi, Federico and Lahijanian, Morteza and Sunberg, Zachary},
  title         = {Robustness Analysis of {POMDP} Policies to Observation Perturbations},
  year          = {2026},
  howpublished  = {arXiv preprint arXiv:2604.21256},
  eprint        = {2604.21256},
  archivePrefix = {arXiv},
  doi           = {10.48550/arXiv.2604.21256},
  url           = {https://arxiv.org/abs/2604.21256}
}

@misc{Mah23,
  author        = {Mahajan, Anuj and Zhang, Amy},
  title         = {Generalization Across Observation Shifts in Reinforcement Learning},
  year          = {2023},
  howpublished  = {arXiv preprint arXiv:2306.04595},
  eprint        = {2306.04595},
  archivePrefix = {arXiv},
  doi           = {10.48550/arXiv.2306.04595},
  url           = {https://arxiv.org/abs/2306.04595}
}

@article{Gar21,
  author  = {Garcia, Damien},
  title   = {{SIMUS}: An Open-Source Simulator for Medical Ultrasound Imaging. {Part I}: Theory and Examples},
  journal = {Computer Methods and Programs in Biomedicine},
  year    = {2022},
  volume  = {218},
  note    = {Art. no. 106726},
  doi     = {10.1016/j.cmpb.2022.106726}
}

@inproceedings{Wan20,
  author    = {Wang, Qiang and Peng, Bo and Cao, Ziyuan and Huang, Xing and Jiang, Jingfeng},
  title     = {A Real-Time Ultrasound Simulator Using Monte-Carlo Path Tracing in Conjunction with {OptiX} Engine},
  booktitle = {2020 IEEE International Conference on Systems, Man, and Cybernetics},
  year      = {2020},
  pages     = {3661--3666},
  doi       = {10.1109/SMC42975.2020.9283057}
}

@inproceedings{Ao25,
  author    = {Ao, Yunke and Moghani, Masoud and Mittal, Mayank and
               Prajapat, Manish and Wu, Luohong and Giraud, Frederic and
               Carrillo, Fabio and Krause, Andreas and F{\"u}rnstahl, Philipp},
  title     = {{SonoGym}: High Performance Simulation for Challenging Surgical Tasks with Robotic Ultrasound},
  booktitle = {Advances in Neural Information Processing Systems},
  volume    = {38},
  year      = {2025},
  note      = {Datasets and Benchmarks Track},
  url       = {https://proceedings.neurips.cc/paper_files/paper/2025/hash/396b1d7ada6d6980d51c81bc5e4da15b-Abstract-Datasets_and_Benchmarks_Track.html}
}

@inproceedings{Son24,
  author    = {Song, Yuhan and Chong, Nak Young},
  title     = {{S-CycleGAN}: Semantic Segmentation Enhanced {CT}-Ultrasound Image-to-Image Translation for Robotic Ultrasonography},
  booktitle = {2024 IEEE International Conference on Cyborg and Bionic Systems},
  year      = {2024},
  pages     = {115--120},
  doi       = {10.1109/CBS61689.2024.10860598}
}

\end{document}